\pdfoutput=1 
\documentclass{article}

\usepackage{arxiv}

\usepackage[utf8]{inputenc} 
\usepackage[T1]{fontenc}    
\usepackage{hyperref}       
\usepackage{url}            
\usepackage{booktabs}       
\usepackage{amsfonts}       
\usepackage{nicefrac}       
\usepackage{microtype}      
\usepackage{lipsum}		
\usepackage{graphicx}
\usepackage{natbib}
\usepackage{doi}

\usepackage{verbatim}
\usepackage{spverbatim}
\DeclareUnicodeCharacter{251C}{\mbox{\kern.23em
  \vrule height2.2exdepth1exwidth.4pt\vrule height2.2ptdepth-1.8ptwidth.23em}}
\DeclareUnicodeCharacter{2500}{\mbox{\vrule height2.2ptdepth-1.8ptwidth.5em}}
\DeclareUnicodeCharacter{2514}{\mbox{\kern.23em
  \vrule height2.2exdepth-1.8ptwidth.4pt\vrule height2.2ptdepth-1.8ptwidth.23em }}
\DeclareUnicodeCharacter{2502}{\mbox{\kern.23em
  \vrule height2.2exdepth-1.8ptwidth.4pt\hspace{1cm}}}
\usepackage{multirow}
\usepackage{longtable}
\usepackage{amsmath}
\usepackage{subcaption}

\usepackage{cleveref}

\title{Machine Learning for Real-World Evidence Analysis of  COVID-19 Pharmacotherapy}

\author{ \href{https://orcid.org/0000-0001-6527-3059}{\includegraphics[scale=0.06]{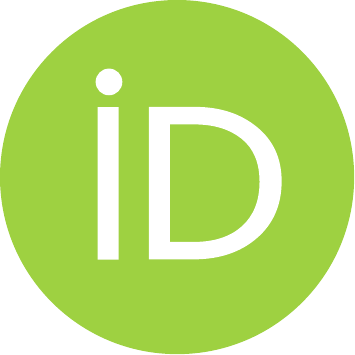}\hspace{1mm}Aurelia Bustos} \\
	AI Medical Research Unit\\
	MedBravo\thanks{MedBravo \url{www.medbravo.org}}\\
	\texttt{aurelia@medbravo.org} \\
	\And 
	\href{https://orcid.org/0000-0002-0547-5975}{\includegraphics[scale=0.06]{orcid.pdf}\hspace{1mm}Patricio Mas\_Serrano} \\
	Pharmacy Department\\
	HGUA\thanks{Hospital General Universitario de Alicante (HGUA)}, ISABIAL\thanks{Institute for Health and Biomedical Research, Alicante, Spain (ISABIAL)} \\
	\texttt{mas\_pat@gva.es} \\
	\And
	Mari Luz Boquera\\
	Pharmacy Department\\
	HGUA\\
	\texttt{mluz\_boquera@gmail.com} \\
	\And
	Jose Maria Salinas\\
	IT Department\thanks{Department of Health Informatics, Hospital Universitario San Juan de Alicante, Spain}\\
	San Juan University Hospital\\
	\texttt{salinas\_josser@gva.es} \\
}

\renewcommand{\shorttitle}{Machine Learning for Real-World Evidence Analysis of  COVID-19 Pharmacotherapy}

\hypersetup{
pdftitle={Machine Learning for Real-World Evidence Analysis of  COVID-19 Pharmacotherapy},
pdfsubject={RWE, AI},
pdfauthor={Aurelia Bustos},
pdfkeywords={RWE, Machine learning, Precision medicine, COVID-19, SARS-CoV-2},
}

\begin{document}

\maketitle
\begin{abstract}
\textbf{Introduction} There is a growing amount of real-world data generated from clinical practice that can be used to analyze the real-world evidence (RWE) of COVID-19 pharmacotherapy and validate the results of randomized clinical trials (RCTs). Machine learning (ML) methods are being increasingly used in RWE and are promising tools for precision-medicine. In this study, ML methods are applied to study retrospectively the efficacy of therapies on COVID-19 hospital admissions in the Valencian Region in Spain.

\textbf{Methods} 5244 and 1312 COVID-19 hospital admissions - dated between January 2020 and January 2021 from 10 health departments, were used respectively for training and validation of separate treatment-effect models (TE-ML) for remdesivir, corticosteroids, tocilizumab, lopinavir-ritonavir, azithromycin and chloroquine/hydroxychloroquine. 2390 admissions from 2 additional health departments were reserved as an independent test to analyze retrospectively the survival benefits of therapies in the population selected by the TE-ML models using cox-proportional hazard models.
TE-ML models were adjusted using treatment propensity scores to control for pre-treatment confounding variables associated to outcome and further evaluated for futility. ML architecture was based on boosted decision-trees.

\textbf{Results} In the populations identified by the TE-ML models, only Remdesivir and Tocilizumab were significantly associated with an increase in survival time, with hazard ratios of 0.41 (P = 0.04) and 0.21 (P = 0.001), respectively. No survival benefits from chloroquine derivatives, lopinavir-ritonavir and azithromycin were demonstrated. Tools to explain the predictions of TE-ML models are explored at patient-level as potential tools for personalized decision making and precision medicine. 

\textbf{Conclusion} ML methods are suitable tools toward RWE analysis of COVID-19 pharmacotherapies. Results obtained reproduce published results on RWE and validate the results from RCTs.

\end{abstract}

\keywords{RWE \and Machine learning \and Precision medicine \and COVID-19 \and SARS-CoV-2}

\section{Introduction}
Evidence from prospective randomized clinical trials (RCTs) on COVID-19 pharmacotherapy efficacy and safety is accumulating and rapidly evolving. RCTs supporting the use of COVID-19 treatments often enroll narrow patient populations. However, a much wider patient population is likely to receive these drugs in clinical practice. As a result there is always a uncertainty between the effectiveness and safety of drugs as obtained from clinical trials and their real benefits and risks in the clinical practice. 
Ambiguity remains about optimal dosing strategies, contraindications for treatment, and the effectiveness of COVID-19 pharmacotherapy used in the treatment of different sub-populations of patients with COVID-19. There is an increasingly amount of real-world data generated from COVID-19 admissions that could potentially be used to understand the clinical practices in COVID-19 management as well as to analyze the real-world evidence (RWE) of COVID-19 pharmacotherapy used. RWE has the potential benefits of being less costly, evaluate patient populations not tipically studied in RCTs and validate the results obtained in RCTs. In spite of the potential benefits of RWE, concerns about the validity and reliability of RWE evaluating medical interventions remain \citep{sherman2016,jarow2017}. Potential disadvantges include lack of controlled and standardized measurements and lack of baseline randomization to control confounding resulting from treatment selection. RWE is helpful for making prescribing decisions only after its limitations are thoroughly addressed and accepted recommendations and methods for quality are followed \citep{franklin2021}.  RWE is a necessary exercise to be done to confirm that the results obtained by clinical trials are replicable in the clinical practice. In medicine, clinical guidelines ensures that drugs are prescribed based on the evidence gathered from clinical studies done during drug development programs as well as post-commercialization programs. Clinical guidelines dictates the indications and populations that would benefit from an intervention and/or drug. Nonetheless even if those recommendations are substantiated by well designed clinical trials, and accumulated evidence, they lack the fine-grained customization required to adapt and tailor the recommendation to a particular patient. One of the most interesting and promising uses of machine learning (ML) in RWE is to build and validate prescription models that would personalize treatment beyond clinical guidelines. ML represents a means toward precision medicine by which the potential effectiveness and safety of specific treatments in a given individual could be predicted \citep{huang2018}.

On this study, we aim to 1) contribute to the validation of RCTs on the effectiveness of most common COVID-19 pharmacotherapy, helping to identify the patient profiles based on their responsiveness to  treatment using readily available data derived from electronic health records (EHRs), 2) follow a methodology widely accepted and successfully applied in a prior study on RWE of COVID-19 therapies \citet{LAM2021} to investigate if their results on the effectiveness of remdesivir and corticosteroids are reproducible in a different population 3) apply the same methods to extend the RWE on other common COVID-19 pharmacotherapy and 4) enrich those experimental approaches in terms of explainability, futility and additive testing and negative treatment effect.

\section{Methods}
\label{sec:methods}
\subsection{Source of data}
After approval by the local EC and regional IRB, iterative extractions covering the period from January 2020 up to January 2021 retrieved the clinical data from hospital admissions of patients attended at 12 health departments from the Valencian Region, who had at least one confirmed positive covid test (serological or RT-PCR test).
Clinical data was anonymized and covered, with varying degree of completeness, patient demography, dates of admissions and discharge, sequences of all hospital services where patients were transferred during each hospitalization, medication, gasometric tests, fluidotherapy, laboratory tests, covid microbiology tests, medical notes from all encounters including emergency room, hospitalizations, and intensive care units in natural language, vital signs, clinical scales, image reports and destination after hospital discharge. 
The source population totaled 12912 patients corresponding to 16718 hospital admissions -referred as \emph{all admissions}-, totaling 5.1 millions data points. Each of the Valencia Region's Health departments 5, 19, 2, 17, 12, 21, 16, 20, 18, 6, 8, 1 contributed a total of  1721, 1303, 1003, 993, 944, 747, 222, 210, 161, 152 and 117 COVID-19 admissions respectively.
\subsection{Data Processing}
\paragraph{Structured and semi-structured data} There were a total of 7680 distinct laboratory name tests and not all of them were normalized to an unequivocal coding system. For cases not normalized, only local codes were available or no coding system was available.  In order to disambiguate the laboratory tests of interest listed in Table~ \ref{tab:Variables2}, the following pre-processing pipeline was applied; if a normalized standard coding system was available it was used and values were converted to the most common unit for each laboratory test. Alternatively, if a normalized standard coding system was not available, then it was assigned to the code of the nearest laboratory test based on the similarity of the text by a fuzzy string matching using the Levenshtein distance, the distance of the medians and the mean Student's t-test comparison between the laboratory value distributions. Finally the obtained clusters were reviewed by a physician and manual regrouping was done when required. 

Missing information was not imputed on this study because of several reasons. First it was not required by the ML methods employed and second, there were concerns of massive imputations of arrays of features with daily values. On the one hand, some of them were either too relevant for disease under study and value should not be imputed, such as temperature. On other cases a missing value had relevant clinical meaning as for example a missing lab test, not done because it was not clinically indicated for a particular clinical case, such as for example Troponines, or because it was already done at a prior time, as this is the case for transaminases that are usually not repeated on a daily basis. 

Diagnoses at discharge were obtained from coded terms using the ICD-10 coding system. Drugs administered were mapped to the WHO Anatomical Therapeutic Chemical (ATC) classification system 5th level.

In-hospital survival information as well as discharge destination was available for all admissions. 

\paragraph{Natural language}
Radiological entities suggestive and or confirmatory of covid pneumonia were extracted from natural language in radiology reports applying a deep-learning multilabel text classifier \citep{Bustos2020} further fine-tuned and validated on spanish radiology reports of x-rays in covid patients \citep{Vaya2020}. Radiological entities and corresponding UMLS Metathesaurus unique identifiers (cuis) for the COVID-19 radiological entities are listed in Table~\ref{tab:CUIs}.
Medical notes from clinical encounters found in natural language were not used on this study.

\subsection{Study Population and Dataset Partitions}
In order to filter only COVID-19 admissions, from \emph{all admissions}, the selection criteria required at least 1 sars-cov-2 RT-PCR positive test or antigen test in a period extending from 3 weeks before the admission up to the discharge date AND either having a COVID-19 coded diagnosis at discharge OR having reported any COVID-19 radiological entity during the admission. The selected radiological entities and corresponding UMLS Metathesaurus cuis for the COVID-19 radiological entities are listed in Table~\ref{tab:CUIs}. Eligible population had to be admitted to hospital, extending more than emergency-ward admission.
After applying the selection criteria, the final study population totaled 8534 patients corresponding to 8971 hospital admissions. 

An independent test set was set aside, consisting of all COVID-19 admissions from two different GVA Health Departments (17 and 19) totaling to a holdout test of 2390 admissions. Reserving entirely two different Health Departments was preferred over other partitioning criteria such as different time periods and/or random criteria so to test the models capability to generalize across different hospitals with varying clinical practices. 
Treatment-effect models (TE-ML) training and validation was done in the set of all other 10 Health departments, randomly choosing a proportion of 80:20 corresponding to 5244 admissions for training and 1312 admissions for validation.

\subsection{Study Variables}
All variables extracted are listed in Tables~\ref{tab:Variables1},~\ref{tab:Variables2} and ~\ref{tab:Variables3}. 
\subsubsection{Composite Variables}
The following composite variables were obtained by combining two or more individual variables as follows:
\paragraph{Modified WHO COVID Outcomes Scale \citep{Cao2020}} The seven-category ordinal scale uses scores 1 to 8 of the WHO-recommended scale but collapses scores 6 and 7 into a single category and renumber the WHO category 8 to 7 (see Table~\ref{tab:who}). As patient status after discharge was not available, the following surrogates were used from the discharge destination to assign WHO categories: 3 if the discharge was a transfer to an acute care hospital ('transfer to acute hospital', 'transfer to another hospital', 'transfer to another acute hospital'), 2 if the discharge was to a chronic care facility and/or home hospitalization unit  ('transfer of residence or assisted socio-sanitary center', 'residence or socio-sanitary center', 'medium and long stay hospital', 'home hospitalization unit', 'medium long stay hospital transfer', 'home care', 'home hospitalization') and 1 if the patient was discharged as an outpatient ('primary care team', 'outpatient consultations', 'address', 'day hospital', 'specialty center', 'voluntary discharge', 'escaped', 'general practitioner', 'midwife',  'mental health unit')
Following recommendations by \cite{Kelly2020} instead of using a limited number of time-point evaluations, a daily trajectory of the score over all days up to discharge was calculated for each patient and included as array of daily values (see Table~\ref{tab:Variables2})

\paragraph{Critical respiratory illness} Critical respiratory illness in the setting of COVID-19 was defined following \cite{Haimovich2020} as any COVID-19 patient meeting one of the following criteria: oxygenation flow rate greater than or equal to 10 L/min, high-flow oxygenation, noninvasive ventilation, invasive ventilation, or death.

\paragraph{Disease severity grade} Disease severity grade was defined following \cite{linssen2020} as G 1-mild (patients admitted to the emergency room but discharged without requiring hospitalization), G 2-moderate (hospitalization <= 5 days without ICU/ventilation, recovered), G 3-severe (prolonged hospitalization > 5 days without ICU/ventilation, recovered), G 4-critical (life threatening requiring intensive care at any stage of hospitalisation, recovered) and G 5-fatal (death). This severity scale was chosen to capture the length of admissions in the outcome measures.

\paragraph{Charlson comorbidity index} The Charlson comorbidity index is a clinical ordinal scale ranging from 0 to 24 from less to more comorbidities that allows an estimation of the expected 10-year survival based on patient medical history and age. Both Charlson index and the estimated 10-year survival were obtained for all patients at the time of admission, regardless of COVID-19 associated respiratory manifestations and severity, following \cite{charlson1987new}

\begin{table}[h!]
    \centering
    \begin{tabular}{c|l}
         Score & Description \\
         \toprule
         1 & not hospitalized with resumption of normal activities \\
         2 & not hospitalized, but unable to resume normal activities \\
         3 & hospitalized, not requiring supplemental oxygen \\
         4 & hospitalized, requiring supplemental oxygen \\
         5 & hospitalized, requiring nasal high-flow oxygen therapy noninvasive mechanical ventilation, or both \\
         6 & hospitalized, requiring ECMO, invasive mechanical ventilation, or both \\
         7 & death \\
    \end{tabular}
    \caption{Modified WHO COVID Outcomes Scale \citep{Cao2020}}
    \label{tab:who}
\end{table}

\subsubsection{Explicative Variables for Treatment Effect ML Models (TE-ML)}
Baseline up to first 24 hours of hospitalization data (including emerging-ward time) were used as independent variables for treatment effect predictions (see section ML Methods \ref{sec:ML}). Explicative variables, or features, included in TE-ML are marked with $^i$ in Tables~\ref{tab:Variables1},\ref{tab:Variables2}  and \ref{tab:Variables3}. Of note, variables that could potentially leak outcome information were excluded from the independent features used to train the models. For variables in the form of arrays only the baseline value and subsequent value in the first 24 hours of hospitalization was used.

\subsubsection{Covariates for Propensity Score Calculation}
\label{sec:covariates_adjust}
To control for confounding by treatment indication (see Section Stabilized Inverse Probability Weighting ~ \ref{sec:sIPTW}), also referred as treatment propensity, it is necessary to include a representative selection of information on admission and patient characteristics. Although there is no consensus as to which covariates to include in the model aimed to predict treatment propensity scores, in most circumstances, it is appropriate to include all measured pretreatment baseline characteristics regardless if they are balanced or not between the two groups. Including posttreatment characteristics in this model should be considered with great caution and should be included only if they are not affected by the treatment. Attending to these criteria, the covariates selected for training the ML models to predict propensity scores for each treatment are marked with $^c$ in Tables~\ref{tab:Variables1},\ref{tab:Variables2} and \ref{tab:Variables3}.

\subsubsection{Treatment Ascertainment for ML Models}
We classified patients as treated or not treated for each of the most common COVID-19 drugs.
COVID-19 drugs selected for the RWE study were those that were administered in at least 500 admissions: azithromycin (n = 4137), Hidroxicloroquine or Cloroquine (n = 1668), systemic corticosteroids -Dexametasone or Metilprednisolone (n = 3861), Lopinavir-ritonavir (n = 1004), Remdesivir (n = 530) and Tocilizumab (n = 829)
Patients were classified as treated if they received each of those treatment by systemic route (IV or PO) within the first 4 days following hospital admission. Data from patients who received these drugs beyond the initial specified treatment windows were excluded from analysis. For each of the selected drugs, we trained a propensity score ML model and an adjusted TE-ML model.

\subsubsection{Outcome Ascertainment}
\label{sec:out-var}
The outcome of interest was survival time (measured in days). Time to in-hospital mortality was known for the entire population. After discharge, survival time was censored up to last available clinical encounter in the follow-up period extending to January 2021. 
TE-ML algorithms to predict treatment responsiveness were trained following \citet{LAM2021} using a binomial logistic  objective that defined a \textit{positive class} (improved disease if
treated vs worsened disease if not treated) and a \textit{negative class} (worsened disease if treated vs improved disease if not treated).  \cite{LAM2021} defined improved disease as a last recorded oxygen saturation of $>=95\%$, or survival (i.e. discharged alive), and
worsened disease was defined as a last recorded oxygen
saturation of $< 95\%$, or death. Nonetheless, last recorded oxygen saturation is directly influenced by supplemental oxygen intervention and hence is not a measure of improved disease. For this reason, we substituted the binarized variable of last recorded saturation of oxygen by a binarized variable of last recorded modified WHO outcomes >= 4 vs < 4 (see Table~\ref{tab:who})

\subsection{Statistics and Machine Learning Methods}
For each of the selected COVID-19 pharmacotherapy, the following pipeline was followed: 
\paragraph{Stabilized Inverse Probability Weighting}
\label{sec:sIPTW}
Inverse probability-of-treatment
weights were used to adjust for confounding.
Inverse-probability weighting counteract confounding by creating a \textit{pseudo-population} in which the treatment is independent of the measured confounders, and consequently, the causal effect of treatment on outcomes could be estimated.
For this end, first, stabilized inverse probability-of-treatment weights (sIPTW), defined as the ratio of the marginal probability of being treated to the propensity score, were calculated for each admission.
For treated patients the sIPTW is given by
\begin{equation}
   w(x) =  \frac{P(T = 1)}{p(x)} = \frac{P(T = 1)}{P(T=1|X=x)}
\end{equation}
and for non-treated patients, the stabilized weight is
\begin{equation}
   w(x) = \frac{1 - P(T = 1)}{p(x)} = \frac{1 - P(T = 1)}{1 - P(T=1|X=x)}
\end{equation}

\noindent where $\boldsymbol{p(x)}$ is the propensity score, $\boldsymbol{P}$ is the probability of being treated with treatment $\boldsymbol{T}$ and $\boldsymbol{X}$ is the set of measured covariates for each admission.
 
Propensity scores for each treatment were calculated by fitting a ML gradient-boosted decision trees model (Catboost) as explained in section \ref{sec:ML}. 

As using original IPTW may artificially increase the total
sample size and furthermore, in cases where individuals have extremely
small propensity scores, IPTW will be large and the estimation of
the treatment effect is then dominated by these few observations
with very large weights. As a result, this can lead to a noticeable increase in the variances of estimated effects. To address this issue and
achieve stabilization in the modeling we redefined according to \citep{xu2010use} the
IPTW as the ratio of the marginal probability of being
treated to the propensity score henceforth, referred to as stabilized IPTW (\textit{sIPTW})

Lastly, each admission was weighted by their corresponding  \textit{sIPTW} creating a \textit{pseudo-population} where the distribution of values for potential co-founders was similar between treated vs non-treated arms. A subset of variables were chosen to visually verify that the variable distributions were similar among both arms in the new \textit{pseudo-population}. sIPTW weighting, obtained from the propensity- score adjustment based on pretreatment confounders, was used for adjusting both the TE-ML model and the time-to-event survival curves.

\paragraph{Machine Learning}
\label{sec:ML}
For both the propensity score ML models and the TE-ML models, the architecture was based on gradient boosted decision trees, implemented using the CatBoost library \citep{catboost} in the Python programming language. During training, a set of decision trees is built consecutively. Each successive tree is built with reduced loss compared to the previous trees. The number of trees is controlled by the starting parameters. 

Propensity scores was calculated for each treatment by fitting a ML CatBoost model with the treatment indicator as the dependent variable and other covariates measured before the treatment assignment as the independent variables (see covariate selection in section \ref{sec:covariates_adjust}). The model was trained and validated in the entire cohort of COVID-19 admissions with proportions 80:20, corresponding to 7176 admissions for training and 1790 for validation. The advantages of using a Catboost model over a multivariate logistic regression model are a) it implicitly handles missing data which is highly prevalent in real-world data extracted from EHR, b) it effectively deals with highly multidimensional data allowing the inclusion of a larger number of covariates than regression methods hence being able to consider a larger number of potential confounders.

A separate TE-ML model for each treatment was trained on the binomial logistic objective (see section \ref{sec:out-var})

10-fold cross-validation was applied with the goal to select model hyper parameters. Training parameters shared across all models trained were learning rate = 0.3, maximum depth = 6, loss function = Cross Entropy, evaluation metric = AUC. The final model was the best model applying early stopping. 

Result metrics for the propensity scores ML models and TE-ML models are shown for validation and independent test set (see Table~ \ref{tab:results}).

Last, we performed and report herein an explainability study of the model predictions using SHAP (SHapley Additive exPlanations)
values\citep{lundberg2017}. SHAP assigns each feature an importance value for a particular prediction. SHAP plots were used to interpret the
clinical properties of each treatment-effect estimator.

\paragraph{Survival Analysis}
\label{sec:Survival_analysis}
Each TE-ML model was applied to the holdout test set
of COVID-19 admissions 24 hours after patient admission. The performance of the sIPTW adjusted TE-ML models was measured using a time-to-event analysis on the test set to explore if patients selected by the ML model were associated with an increase in survival time. Survival time was measured through adjusted hazard ratios (HRs). The same method for adjustment for confounding applied to the TE-ML models, was applied to the survival analysis, adjusting also for the covariates described in section \ref{sec:covariates_adjust}. In essence, each admission was weighted by the IPTWs in the time-to-event models. To mitigate the effects of any misspecification in a
model in the IPTWs, all adjustment covariates were also included in the final time-to-event models.
A comparison of the survival time of treated and non treated patients was done 1) in the full population, 2) in the population of test patients who received supplemental oxygen (a more critically ill population, and a population for
whom COVID-19 drugs are explicitly recommended per current clinical guidelines \citep{NIH}) and 3) in the population of test patients predicted to derive a positive effect by the sIPTW adjusted TE-ML model and 4) in the population of test patients predicted to derive non or negative effect by the sIPTW adjusted TE-ML model. 
Analyses were performed, and are presented,
separately for each COVID-19 treatment. We examined the associations between each
treatment and mortality in unadjusted models (eg,
models containing neither adjustment covariates nor
IPTWs) and adjusted time-to-event models. For all
analyses, the level of significance was set at $\alpha = 0.05$.

\paragraph{Futility Analysis}
\label{sec:futility}
Because the present study utilized data from a cohort in which treatment was not randomized, it is possible that residual confounding still influence the results despite the efforts to adjust for confounding variables. For this analysis we assume the worst case scenario where residual confounding always still exists after adjustment and hence not only the intervention (therapy administration or not) but also confounding variables are determining the patient outcomes. If an adjusted TE-ML model is capable of identifying any treatment signal of efficacy over confounding, then there should be an additive effect on the strength of survival association for patients selected with this model vs a dummy model in the training set. Conversely if applying the dummy model does not deteriorate the strength of the association achieved with the adjusted TE-ML model then we conclude that the signal of the treatment efficacy could not be detected by the model and observed differences in outcome were likely attributed to confounding instead of intervention. In order to avoid introducing additional uncertainties both the futile model training and survival analysis testing was done in the training set. Following this hypothesis, for each of the therapies where the TE-ML model resulted in the test set in statistically significant survival gains we did the following futility experiment: First an adjusted TE-ML model was trained with the binomial outcome substituted with randomly assigned values 0 or 1. The ML models, trained on this manner for each therapy, consistently yielded an AUC <= 0.5 in the validation set and hence were considered futile models. Second, we compared the survival results on the selected population in the training set of the adjusted ML futile models vs those from the adjusted treatment-effect models. Only if the strength of survival associations deteriorated, comparing the HR, 95\% CI and p-values, in the population selected by adjusted futile ML model, the adjustment of confounding variables were considered valid enough to allow the model to detect a treatment effect signal, otherwise further experiments were terminated and no conclusions on survival effect were drawn for the intervention. 

\section{Results}
\subsection{Patient Population Description}
\begin{figure*}[h!]
\centering
\includegraphics[width=0.49\linewidth]{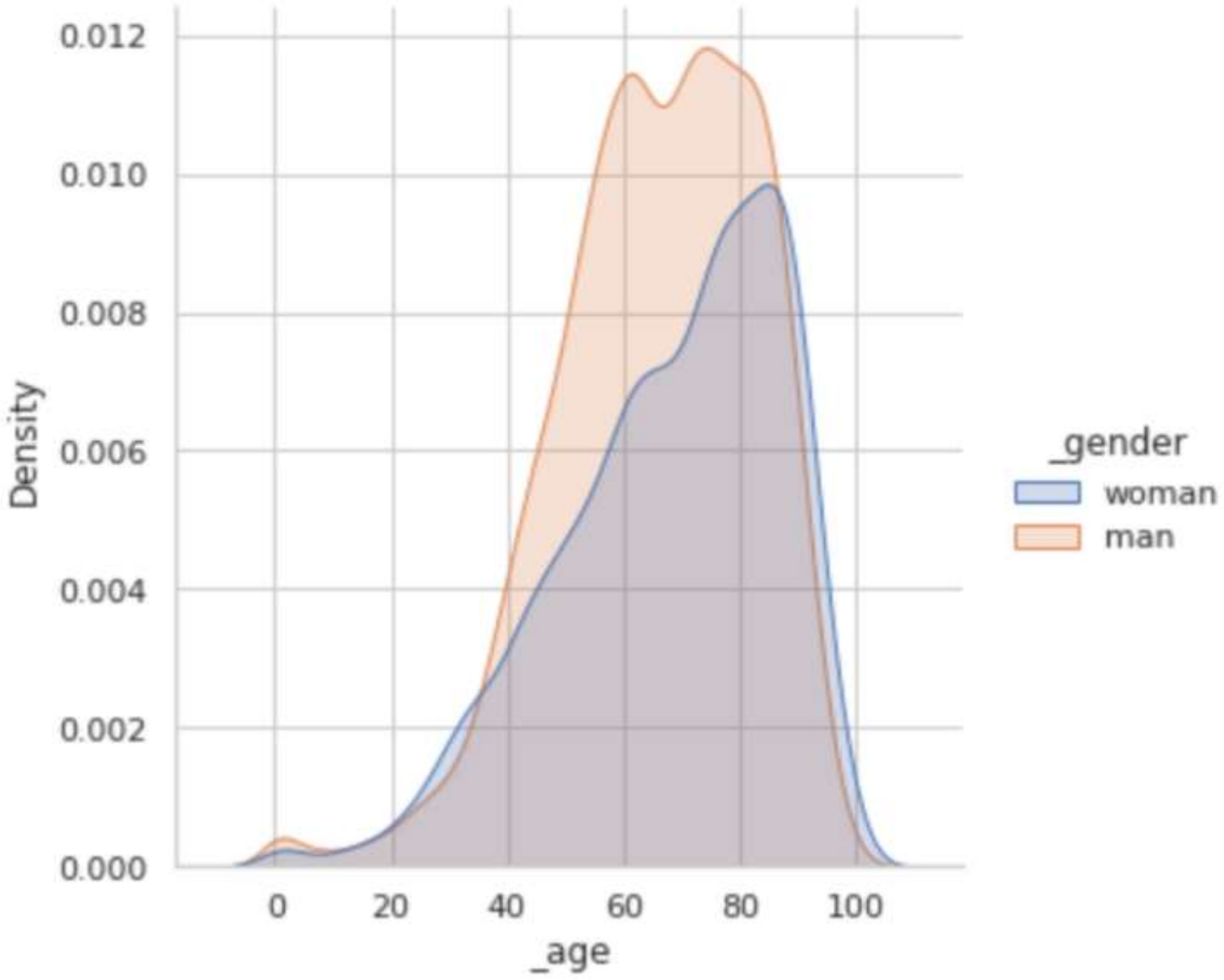}
\caption{Age distribution of COVID-19 admissions by gender}
\label{fig:Age}
\end{figure*}

\begin{figure*}[h!]
\centering
\includegraphics[width=0.49\linewidth]{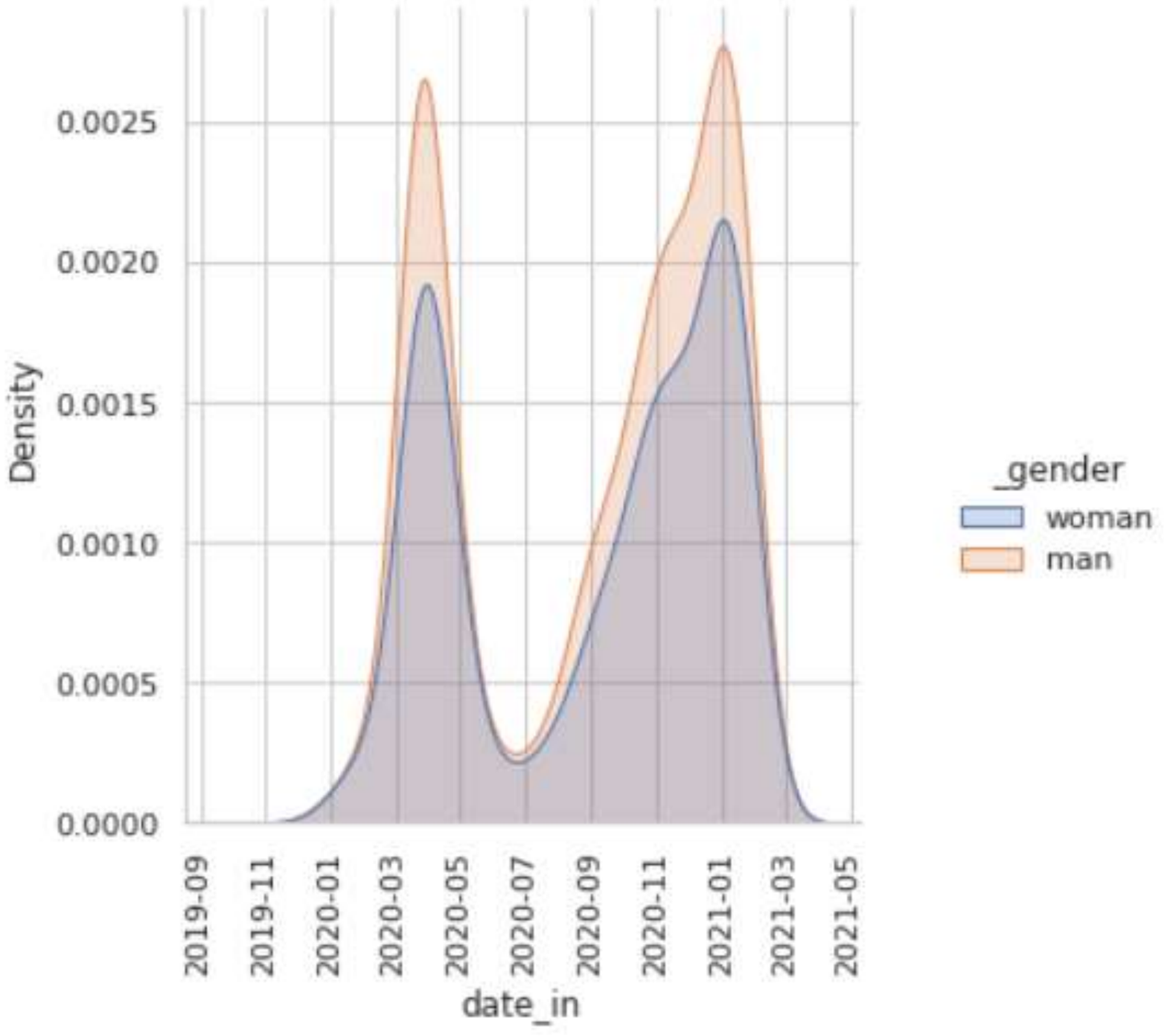}
\caption{Distribution of COVID-19 admission dates by gender}
\label{fig:Date}
\end{figure*}

For COVID-19 admissions (n = 8791), the mean age of patients was 66.5 +/-18.2 (0.0, 101.0) years and 56.1\% were male. The mean age of hospitalized men were 65.4 +/-17.3 (0.0, 101.0) vs 67.8 +/-19.2 (0.0, 101.0) years in women (see age distribution in Fig~\ref{fig:Age}). Mortality was higher for men 17.0\% vs women 14.3\%, being the mean age at death of 78.9 +/-10.7 (31.0, 101.0) and 83.0 +/-10.5 (30.0, 101.0) years respectively. There were a total of 979(10.9\%) ICU admissions with a length of 14 +/-15 (0 , 209) days vs 4 +/-6 (0 , 56) days for non-COVID-19 ICU admissions. The mean hospitalization length of COVID-19 admissions was 10 +/-10 (0 , 247) days vs 5 +/-7 (0 , 145) days for non-COVID-19 related admissions The distribution of dates of admissions is shown in Fig~\ref{fig:Date}

\subsection{Performance Metrics of ML Models}
Table~\ref{tab:results} shows for each of the selected COVID-19 therapies, the performance achieved by the propensity score predictive model and the treatment effect predictive model on their respective final tasks, treatment prescription likelihood (yes vs no) and treatment response (positive vs negative classes).
Models trained to predict for each therapy the prescription likelihood, achieved a high performance for lopinavir-ritonavir, chloroquine or hydroxychloroquine, remdesivir and tocilizumab, being all of those AUCs above 0.90 in the validation set. As described in section \ref{sec:sIPTW}, these models were used to estimate the propensity scores. 
Models trained to predict the treatment response, after adjusted by \textit{sIPTW} achieved the highest AUCs for Tocilizumab, Remdesivir and Lopinavir-Ritonavir ranging from 0.82 to 0.83 in the validation set and their predictive performance degraded to 0.73-0.74 in the out-held test set.

\begin{table}[h!]
    \centering
    \begin{tabular}{p{0.2\textwidth}p{0.15\textwidth}p{0.15\textwidth}p{0.15\textwidth}p{0.15\textwidth}p{0.1\textwidth}}
    \toprule
     \textbf{Model}& \textbf{Propensity-Score} &\textbf{Adjusted Dummy} &\textbf{Adjusted Treatment-Effect} &\\
     Final Task &Treatment prescription likelihood (yes vs no) &Dummy outcome (positive vs negative) &Treatment response (positive vs negative)&\\
     \cmidrule(r){4-5}
     \textbf{Metric} & \textbf{AUC Validation\newline(N = 1790)} &\textbf{AUC Validation\newline(N = 1312)}&\textbf{AUC Validation\newline(N = 1312)}&\textbf{AUC Test\newline(N = 2390)}\\
     \midrule
    \textbf{COVID-19 Therapy} & & &  &  \\ 
    Corticosteroids &0.88 & N/A& 0.75 & 0.55 \\ 
    Remdesivir & 0.92& 0.47 &0.83*  & 0.73 \\ 
    Chloroquine & 0.96 &N/A& 0.75 & 0.66 \\ 
    Azithromycin & 0.84 &N/A& 0.68 &0.51 \\ 
    Lopinavir-Ritonavir & 0.97& 0.5&0.82*&0.74 \\
    Tocilizumab   &0.90 & 0.48 &0.83*&0.73 \\
    \bottomrule
    \end{tabular}
    \caption{Machine learning models: This table summarizes all model trained, their final tasks, metrics and results. Propensity score models were trained and validated in the entire cohort of COVID-19 admissions as an independent test was not required. On the contrary all adjusted models for treatment effect and futile study were trained and validated in COVID-19 admissions excluding health departments 17 and 19 reserved for testing. Propensity adjusted dummy models for the futile analysis were only trained for therapies where the adjusted TE-ML models resulted in statistically significant survival benefits in the test set (marked with *). Legend: * = Adjusted TE-ML models where the survival analysis yielded statistically significant results in the test set. N/A = Not Applicable }
    \label{tab:results}
\end{table}

\subsection{Survival Analysis Testing of the Treatment Effect Models}
\begin{figure}[h!]
\centering
\begin{subfigure}{.49\linewidth}
  \centering
  \label{fig:NonAdjustedGeneralCovidPopulationCort}
  \caption{Non-adjusted General Covid Population}
  \includegraphics[width=1\linewidth]{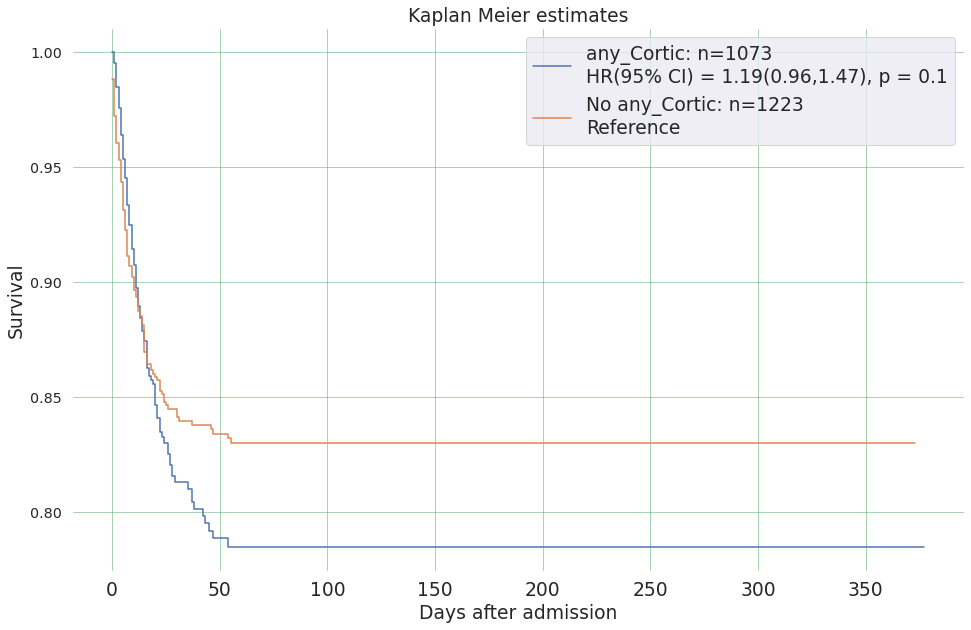}
\end{subfigure}
\begin{subfigure}{.49\linewidth}
  \centering
  \label{fig:AdjustedGeneralCovidPopulationCort}
  \caption{Adjusted General Covid Population} 
  \includegraphics[width=1\linewidth]{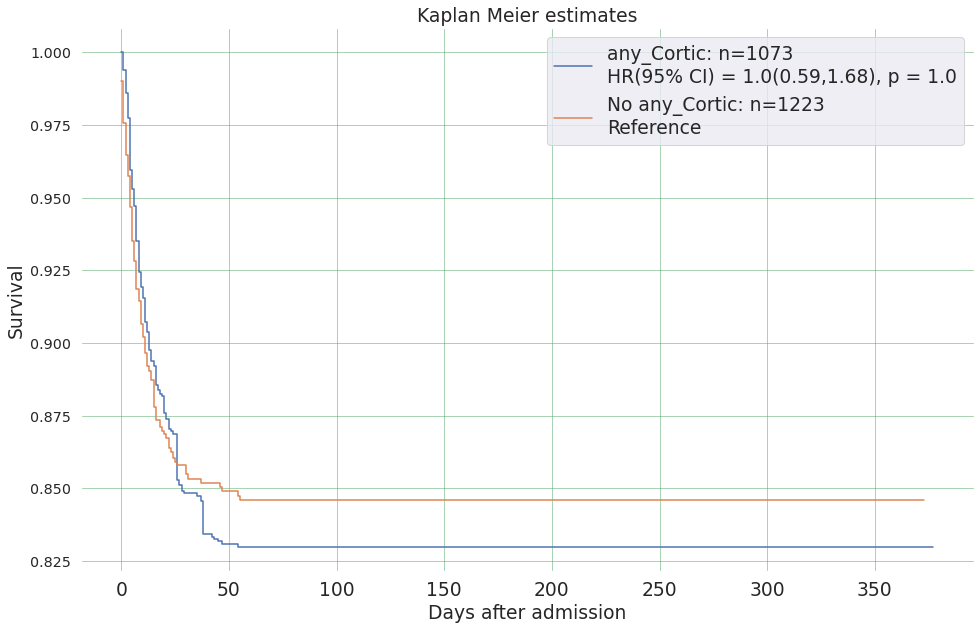}
\end{subfigure}
\begin{subfigure}{.49\textwidth}
  \centering
  \label{fig:AdjustedGuidelinesIndicatedCovidPopulationCort}
  \caption{Adjusted Guidelines Indicated} 
  \includegraphics[width=1\linewidth]{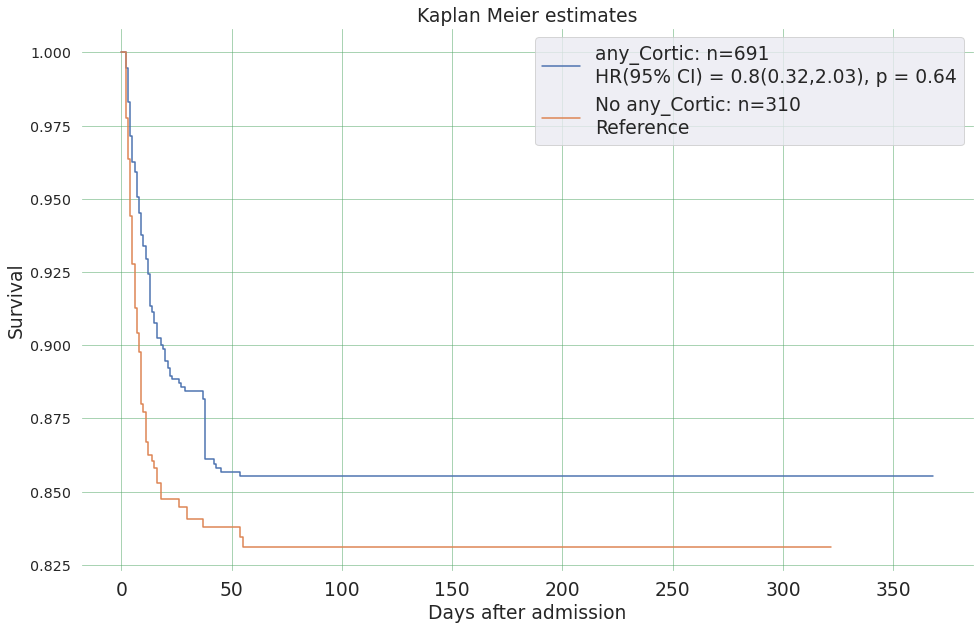}
\end{subfigure}
\begin{subfigure}{.49\textwidth}
  \centering
  \label{fig:AdjustedMLIndicatedCovidPopulationCort}
  \caption{Adjusted ML Indicated} 
  \includegraphics[width=1\linewidth]{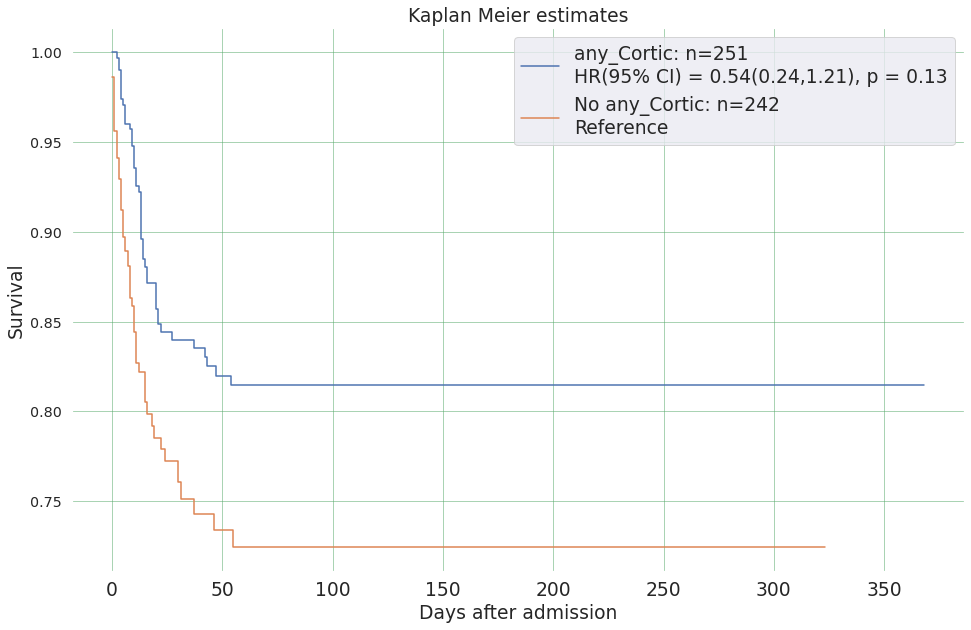}
\end{subfigure}
\begin{subfigure}{.49\textwidth}
  \centering
  \label{fig:AdjustedMLNoIndicatedCovidPopulationCort}
  \caption{Negative Treatment Effect: Adjusted ML Non-Indicated} 
  \includegraphics[width=1\linewidth]{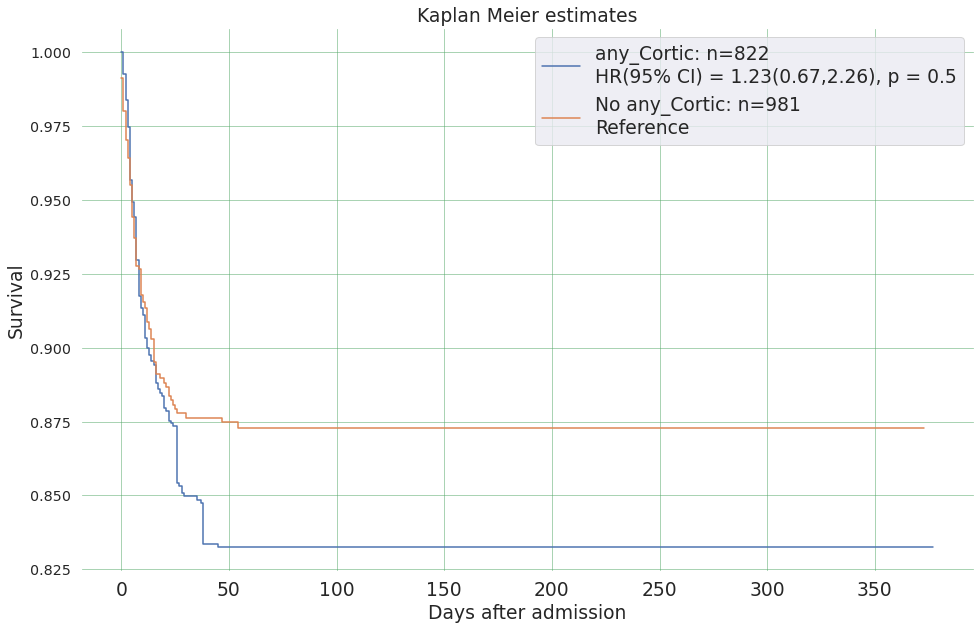}
\end{subfigure}

\caption{\textbf{Corticosteroids}: Survival analysis in the test population.a) Non-adjusted survival curves in treated vs non treated test population.  b) Adjusted survival curves in treated vs non treated test population. c) Adjusted survival curves in treated vs non treated test population who were indicated for treatment by clinical guidelines. d) Adjusted survival curves in treated vs non treated test population who were indicated by the ML model. e) Adjusted survival curves in treated vs non treated test population who were not indicated for treatment by the ML model.}
\label{fig:Cort}
\end{figure}

The performance of the treatment-effect models were tested in 2390 admissions of unique patients, of whom 1073 received systemic corticosteroids, 255 received remdesivir, 664 received azithromycin, 384 received hydroxycholoroquine or chloroquine, 142 received lopinavir-ritonavir and 230 received tocilizumab. 
\paragraph{Systemic corticosteroids -dexametasone and metilprednisolone}
As shown in Fig~\ref{fig:Cort} in the unadjusted time-to-event analysis, corticosteroid
use was not associated with a decrease in survival time in the general population (HR = 1.19;
P = 0.1). After adjustment for confounding by indication, corticosteroid use was not associated
with survival in the COVID-19 test population. Among patients requiring supplemental oxygen (N = 1001) and those indicated by the ML model (N = 493), the relationship remained statistically non-significant,
although the point estimate supported a survival
benefit (HR = 0.8; P = 0.64) which was more pronounced for patients indicated by the TE-ML model (HR = 0.54; P = 0.13). Moreover, this protective treatment disappeared and the hazard inverted as shown in the survival analysis in the group of patients non-indicated to receive corticostisteroids by the ML-models (HR = 1.2; p = 0.5)

\paragraph{Remdesivir}
The unadjusted time-to-event analysis in the unselected test population found that
remdesivir use was associated with a non-significantly increase in survival time (HR = 0.7; P = 0.07). Adjustment for confounding attenuated the relationship and the protective effect disappeared. The adjusted association in
the group that received supplemental oxygen was similarly nonsignificant. However, remdesivir use was statistically significant associated with an increase in survival among those indicated by the algorithm (N = 319) as
suitable for treatment with remdesivir (HR = 0.41;
P = 0.04). This protective effect completely disappeared and even was inverted as shown in the survival analysis in the group of patients non-indicated to receive remdesivir by the ML-models (HR = 1.4; p = 0.2). 
As explained in the futility analysis test (see section \ref{sec:methods}), a propensity adjusted ML futile model was trained in a dummy outcome reaching an AUC = 0.47. The survival benefit was not longer statistically significant when the futile model was applied. In essence, differences in survival gains were not significant for the patient population selected by the futile model who received the treatment vs those selected by the futile model who did not receive it. This supports that the adjustment of confounding variables was adequate to allow the ML to detect a treatment effect signal over the effects of confounding factors and hence further experiments on survival analysis could be trusted
These results support that the TE-ML model can identify patients in whom remdesivir use is associated with improved
survival outcomes and conversely can identify those that would not derive any survival benefit. 
\begin{figure}[h!]
\centering
\begin{subfigure}{.49\linewidth}
  \centering
  \label{fig:NonAdjustedGeneralCovidPopulationRemdesivir}
  \caption{Non-adjusted General Covid Population}
  \includegraphics[width=1\linewidth]{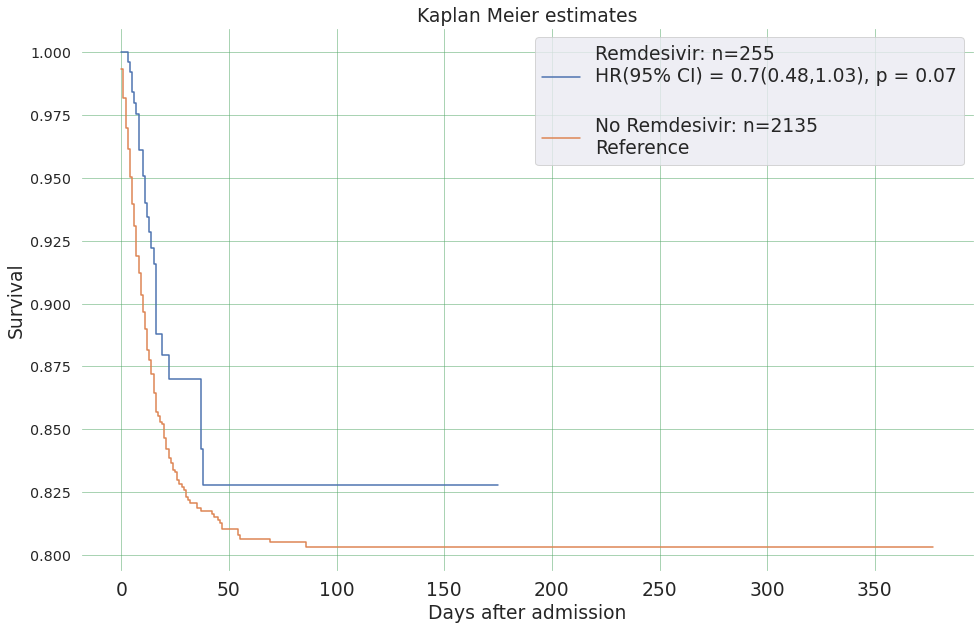}
\end{subfigure}
\begin{subfigure}{.49\linewidth}
  \centering
  \label{fig:AdjustedGeneralCovidPopulationRemdesivir}
  \caption{Adjusted General Covid Population} 
  \includegraphics[width=1\linewidth]{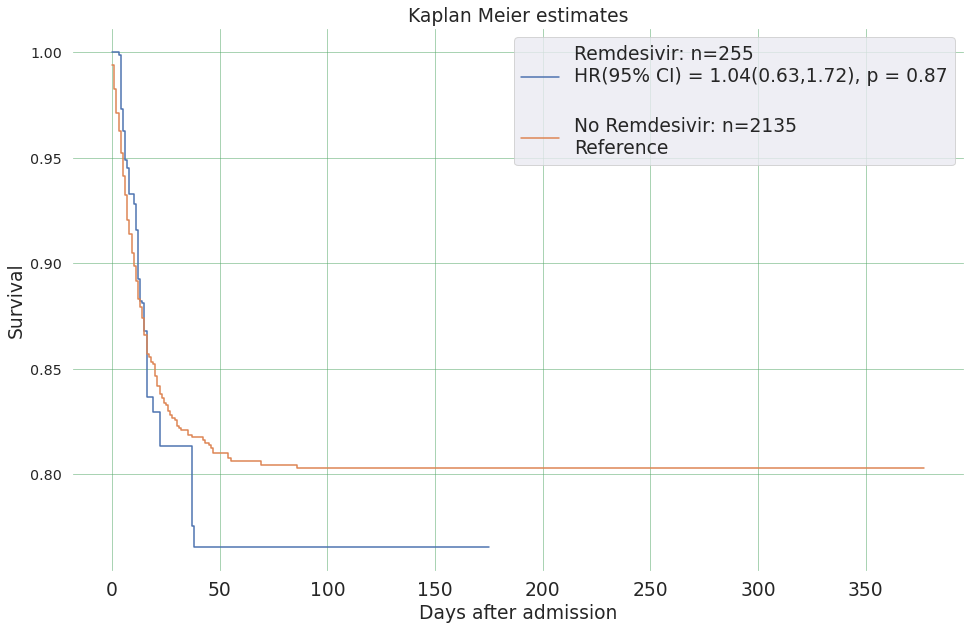}
\end{subfigure}
\begin{subfigure}{.49\textwidth}
  \centering
  \label{fig:AdjustedGuidelinesIndicatedCovidPopulationRemdesivir}
  \caption{Adjusted Guidelines Indicated} 
  \includegraphics[width=1\linewidth]{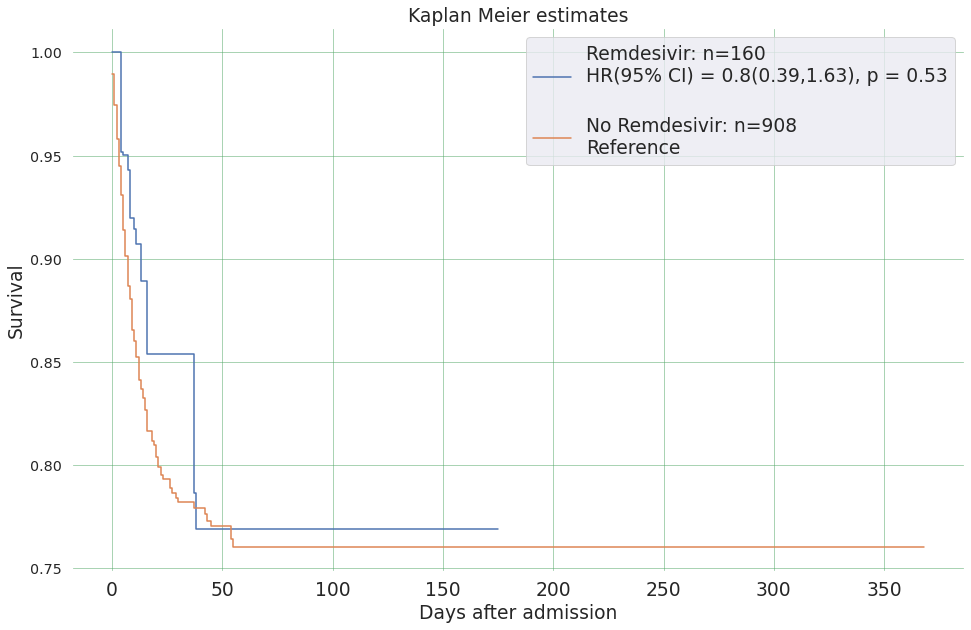}
\end{subfigure}
\begin{subfigure}{.49\textwidth}
  \centering
  \label{fig:AdjustedMLIndicatedCovidPopulationRemdesivir}
  \caption{Adjusted ML Indicated} 
  \includegraphics[width=1\linewidth]{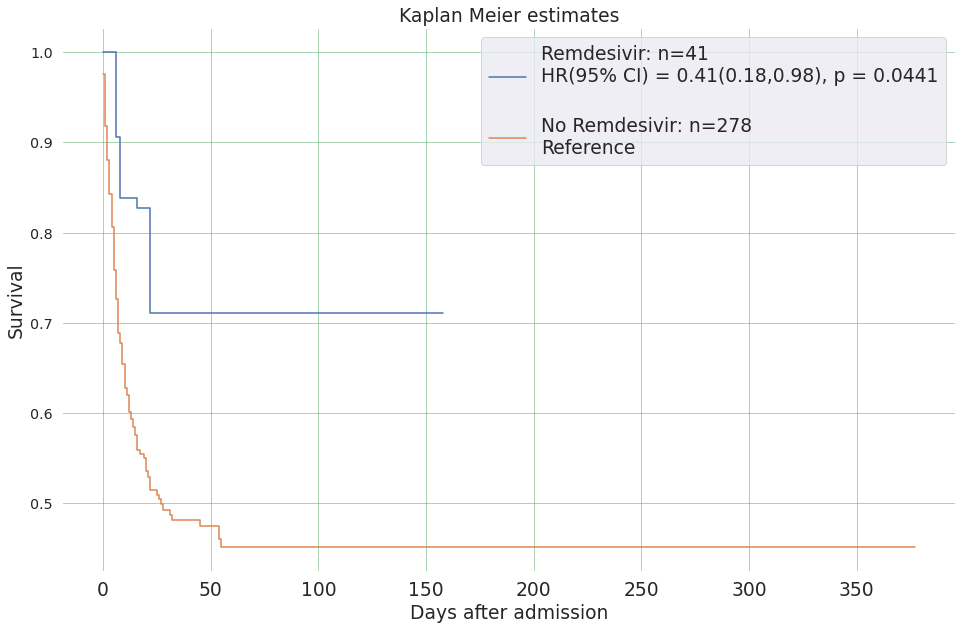}
\end{subfigure}
\begin{subfigure}{.49\textwidth}
  \centering
  \label{fig:AdjustedMLNoIndicatedCovidPopulationRemdesivir}
  \caption{Negative Treatment Effect: Adjusted ML Non-Indicated} 
  \includegraphics[width=1\linewidth]{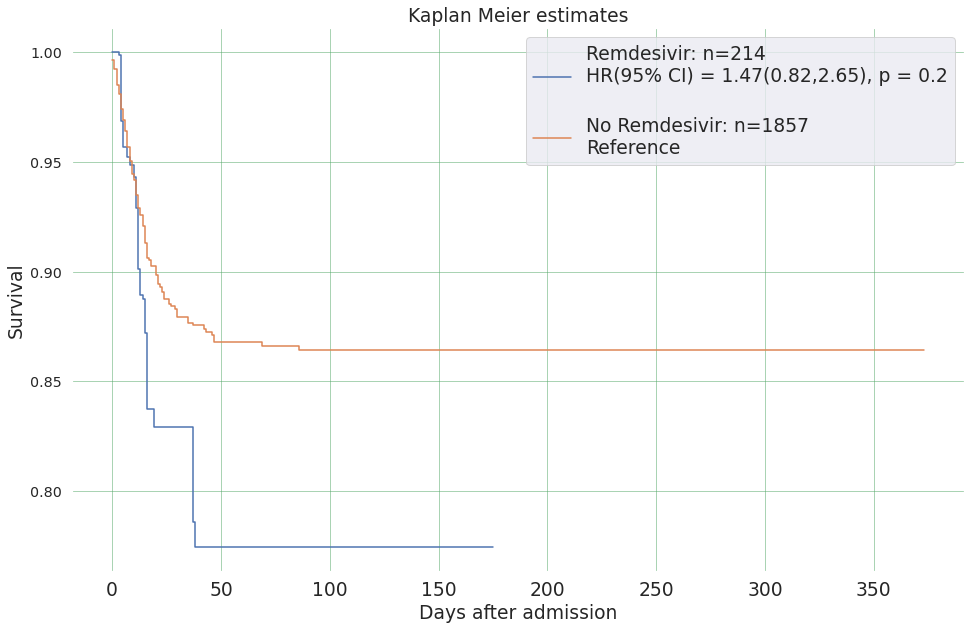}
\end{subfigure}

\caption{\textbf{Remdesivir}: Survival analysis in the test population.a) Non-adjusted survival curves in treated vs non treated test population.  b) Adjusted survival curves in treated vs non treated test population. c) Adjusted survival curves in treated vs non treated test population who were indicated for treatment by clinical guidelines. d) Adjusted survival curves in treated vs non treated test population who were indicated by the ML model. e) Adjusted survival curves in treated vs non treated test population who were not indicated for treatment by the ML model.}
\label{fig:Remdesivir}
\end{figure}

\begin{figure}[h!]
\centering
\begin{subfigure}{.49\linewidth}
  \centering
  \label{fig:NonAdjustedGeneralCovidPopulationAzitromicine}
  \caption{Non-adjusted General Covid Population}
  \includegraphics[width=1\linewidth]{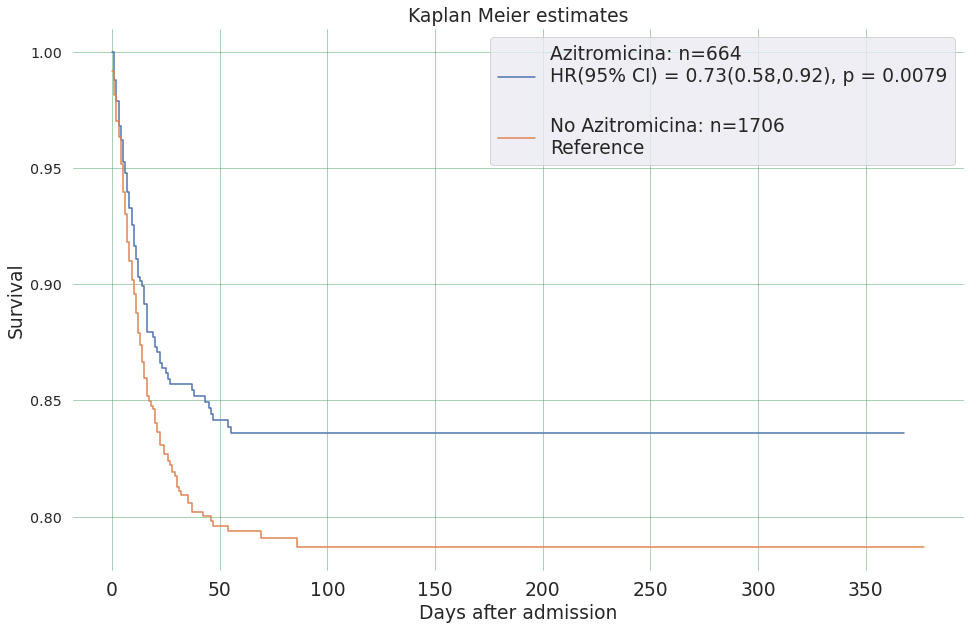}
\end{subfigure}
\begin{subfigure}{.49\linewidth}
  \centering
  \label{fig:AdjustedGeneralCovidPopulationAzitromicine}
  \caption{Adjusted General Covid Population} 
  \includegraphics[width=1\linewidth]{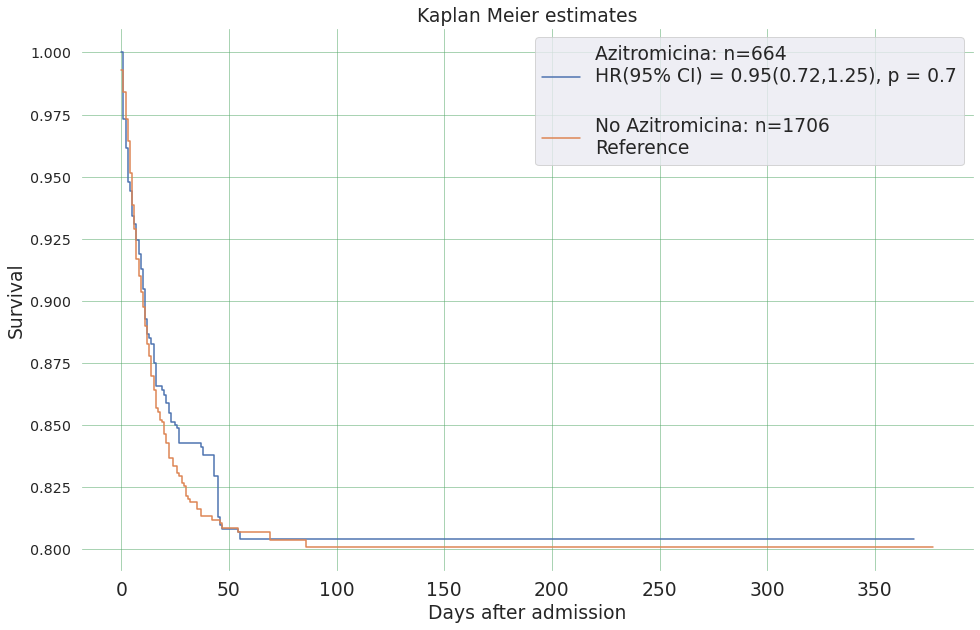}
\end{subfigure}
\begin{subfigure}{.49\textwidth}
  \centering
  \label{fig:AdjustedGuidelinesIndicatedCovidPopulationAzitromicine}
  \caption{Adjusted Guidelines Indicated} 
  \includegraphics[width=1\linewidth]{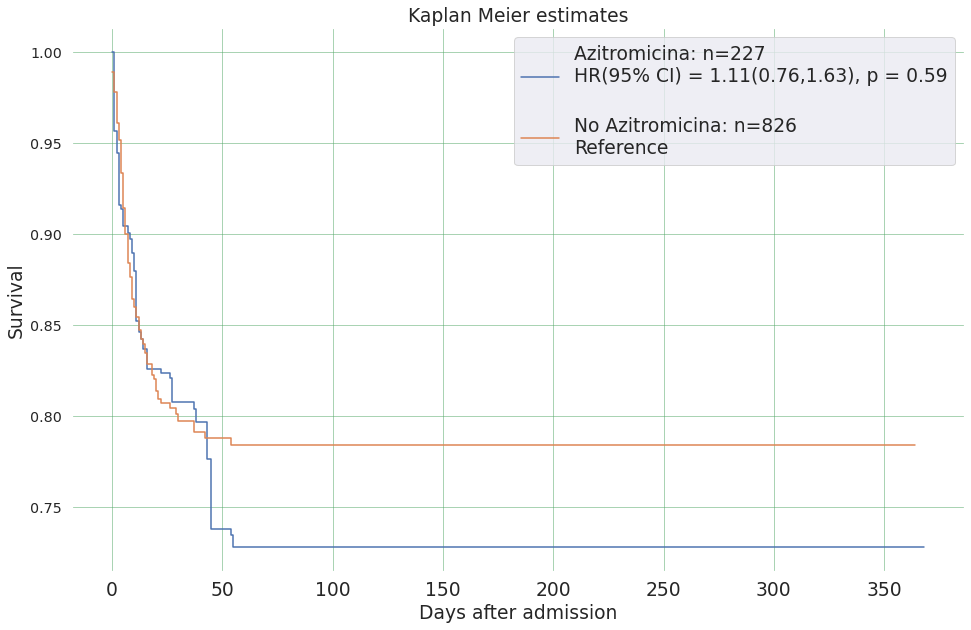}
\end{subfigure}
\begin{subfigure}{.49\textwidth}
  \centering
  \label{fig:AdjustedMLIndicatedCovidPopulationAzitromicine}
  \caption{Adjusted ML Indicated} 
  \includegraphics[width=1\linewidth]{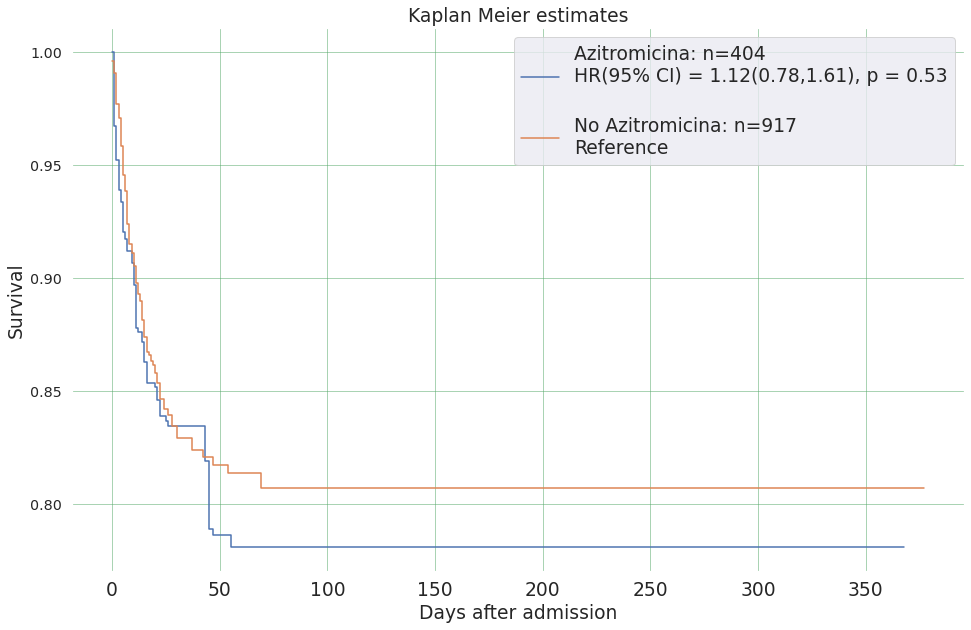}
\end{subfigure}

\caption{\textbf{Azithromycin}: Survival analysis in the test population.a) Non-adjusted survival curves in treated vs non treated test population.  b) Adjusted survival curves in treated vs non treated test population. c) Adjusted survival curves in treated vs non treated test population who were indicated for treatment by clinical guidelines. d) Adjusted survival curves in treated vs non treated test population who were indicated by the ML model.}
\label{fig:Azitromicine}
\end{figure}

\paragraph{Azithromycin} As shown in Fig~\ref{fig:Azitromicine} in the unadjusted time-to-event analysis, azithromycin
use was associated with an increase in
survival time in the general population (HR = 0.73;
P = 0.0079). After adjustment for confounding by
indication, azithromycin use was not associated
with survival in the unselected test of COVID-19 population. Among patients requiring supplemental
oxygen and as well as those indicated by the ML model, the relationship remained statistically nonsignificant,
and the point estimate did not support any survival benefit, being both hazard ratios above 1. These
results support that there is no benefit on terms of survival neither in the unselected COVID-19 population after adjusting for confounding factors related to treatment prescription, not in the selected population requiring supplemental oxygen, not in the patients indicated by the adjusted ML model.

\begin{figure}[h!]
\centering
\begin{subfigure}{.49\linewidth}
  \centering
  \label{fig:NonAdjustedGeneralCovidPopulationCQ}
  \caption{Non-adjusted General Covid Population}
  \includegraphics[width=1\linewidth]{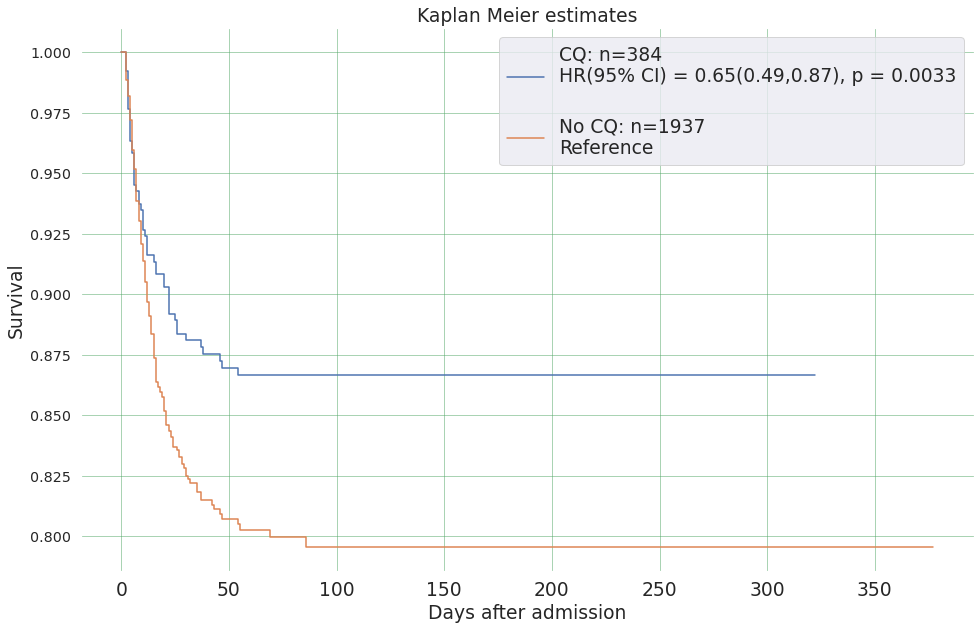}
\end{subfigure}
\begin{subfigure}{.49\linewidth}
  \centering
  \label{fig:AdjustedGeneralCovidPopulationCQ}
  \caption{Adjusted General Covid Population} 
  \includegraphics[width=1\linewidth]{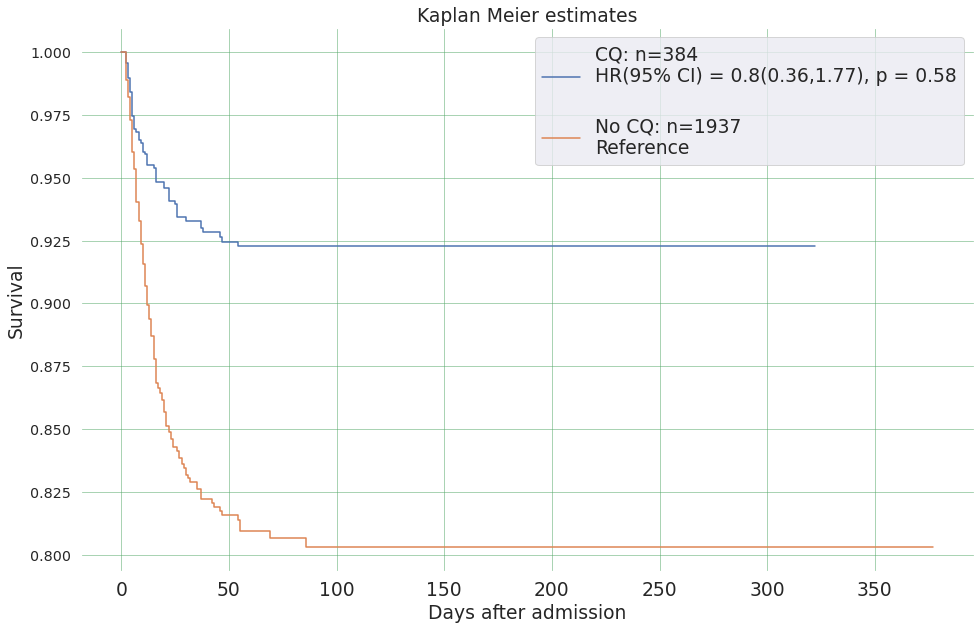}
\end{subfigure}
\begin{subfigure}{.49\textwidth}
  \centering
  \label{fig:AdjustedGuidelinesIndicatedCovidPopulationCQ}
  \caption{Adjusted Guidelines Indicated} 
  \includegraphics[width=1\linewidth]{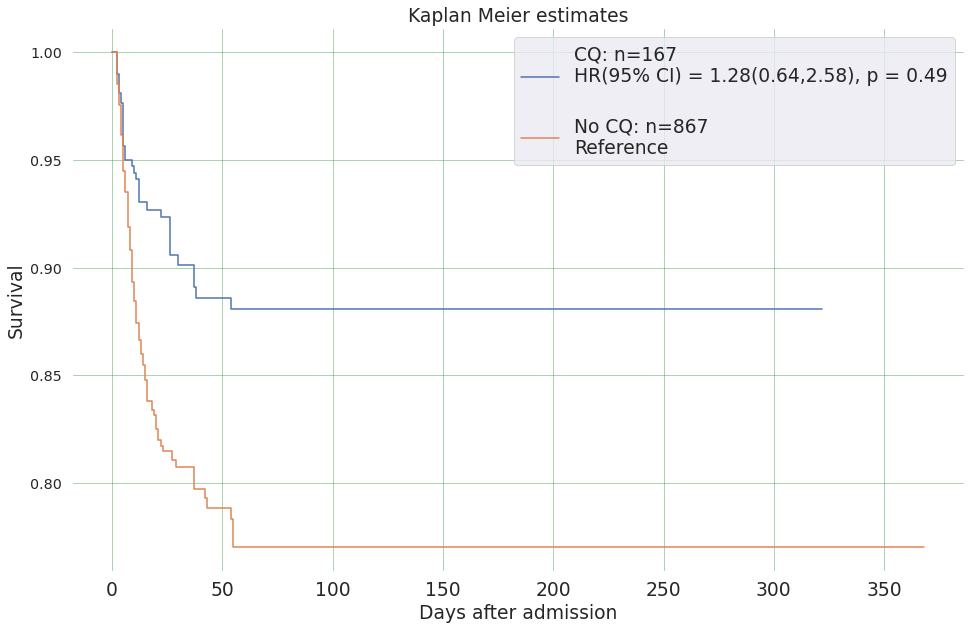}
\end{subfigure}
\begin{subfigure}{.49\textwidth}
  \centering
  \label{fig:AdjustedMLIndicatedCovidPopulationCQ}
  \caption{Adjusted ML Indicated} 
  \includegraphics[width=1\linewidth]{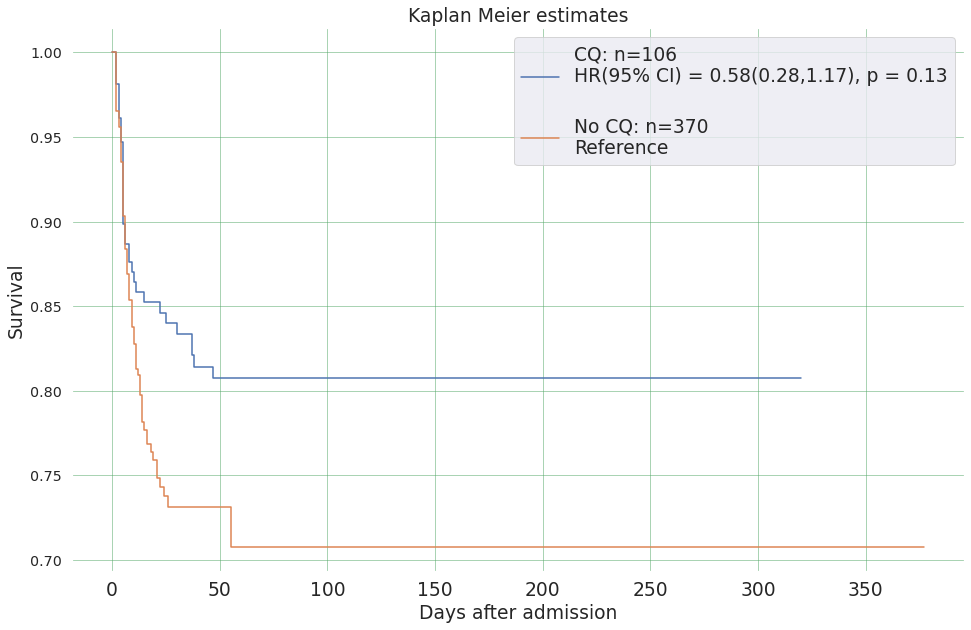}
\end{subfigure}
\caption{\textbf{Hidroxicloroquine or Cloroquine}: Survival analysis in the test population.a) Non-adjusted survival curves in treated vs non treated test population.  b) Adjusted survival curves in treated vs non treated test population. c) Adjusted survival curves in treated vs non treated test population who were indicated for treatment by clinical guidelines. d) Adjusted survival curves in treated vs non treated test population who were indicated by the ML model. }
\label{fig:CQ}
\end{figure}

\begin{figure}[h!]
\centering
\begin{subfigure}{.49\linewidth}
  \centering
  \label{fig:NonAdjustedGeneralCovidPopulationLopinavir_Ritonavir}
  \caption{Non-adjusted General Covid Population}
  \includegraphics[width=1\linewidth]{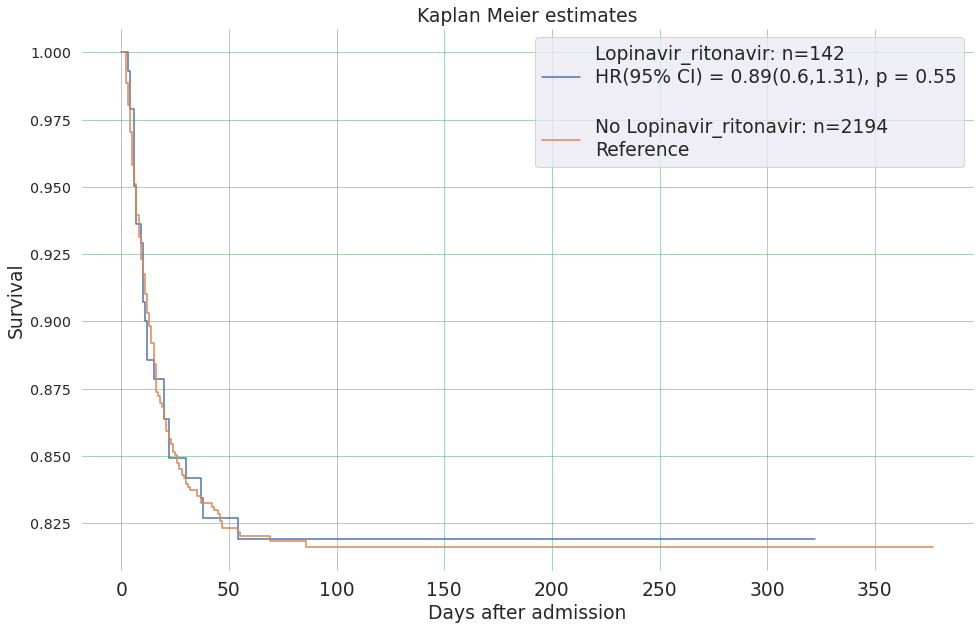}
\end{subfigure}
\begin{subfigure}{.49\linewidth}
  \centering
  \label{fig:AdjustedGeneralCovidPopulationLopinavir_Ritonavir}
  \caption{Adjusted General Covid Population} 
  \includegraphics[width=1\linewidth]{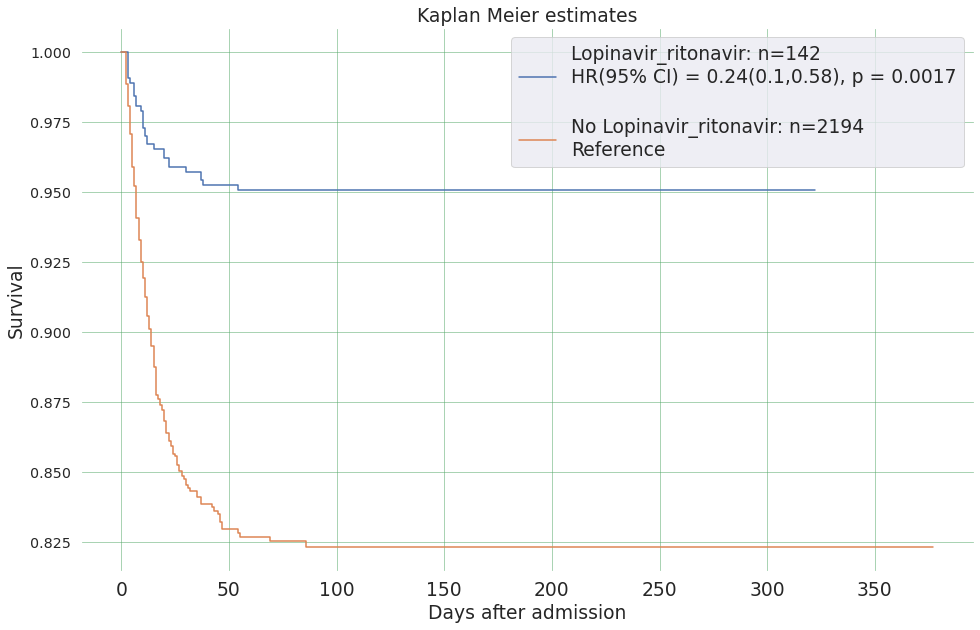}
\end{subfigure}
\begin{subfigure}{.49\textwidth}
  \centering
  \label{fig:AdjustedGuidelinesIndicatedCovidPopulationLopinavir_Ritonavir}
  \caption{Adjusted Guidelines Indicated} 
  \includegraphics[width=1\linewidth]{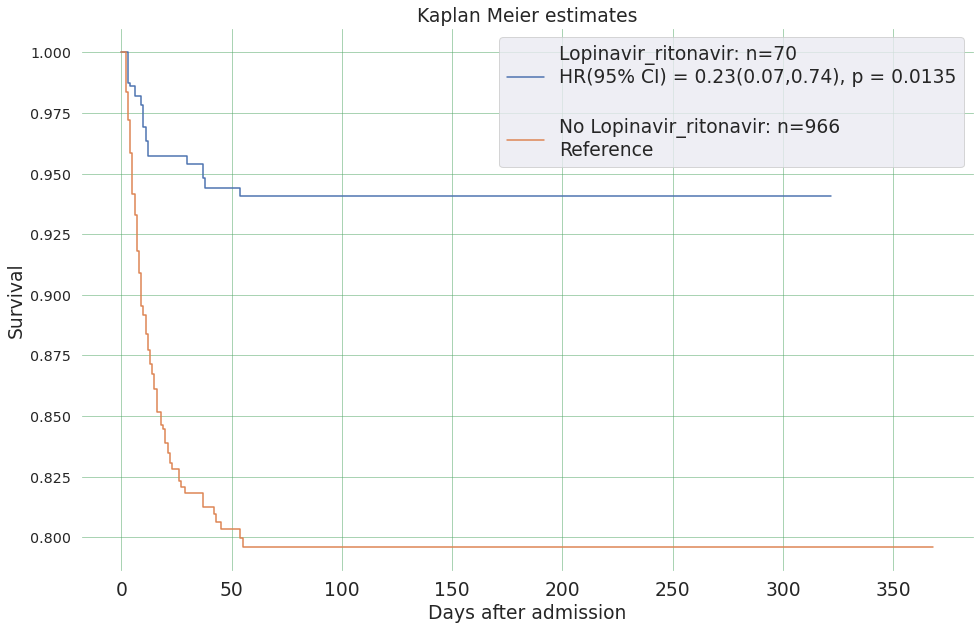}
\end{subfigure}
\caption{\textbf{Lopinavir-Ritonavir}: Survival analysis in the test population.a) Non-adjusted survival curves in treated vs non treated test population.  b) Adjusted survival curves in treated vs non treated test population. c) Adjusted survival curves in treated vs non treated test population who were indicated for treatment by clinical guidelines.}
\label{fig:Lopinavir-Ritonavir}
\end{figure}

\paragraph{Hidroxicloroquine or Cloroquine} Similarly to the results for azithromycin, as shown in Fig~\ref{fig:CQ}, in the unadjusted time-to-event analysis, hydroxychloroquine or chloroquine use was associated with an increase in
survival time in the general unselected population (HR = 0.65;
P = 0.0033). After adjustment for confounding by indication the association was not longer statistically significant neither in the unselected population (HR = 0.8, P = 0.58), not among patients requiring supplemental oxygen (HR = 1.2, P = 0.49) and not among those indicated by the ML model (HR = 0.58, P = 0.13). These
results support that there is no benefit on terms of survival from the use of Hidroxicloroquine or Cloroquine.
\paragraph{Lopinavir-ritonavir}
As shown in Fig~\ref{fig:Lopinavir-Ritonavir}, the unadjusted time-to-event analysis did not find any association of 
lopinavir-ritonavir use with survival time, and survival curves of both treated and not treated patients overlapped.
Adjustment for confounding revealed a relationship,
where lopinavir-ritonavir was significantly associated an increase in survival time in the general unselected population (HR = 0.24 , P = 0.0017). The adjusted survival association in
the group that received supplemental oxygen as well as in the group indicated by the TE-ML model were
similarly significant, yielding a HR = 0.23 ( P = 0.00135) and a HR = 0.21 ( P = 0.00115) respectively. As explained in the futility analysis test \ref{sec:methods}, a propensity adjusted ML futile model was trained in a dummy outcome reaching an AUC = 0.5 and this futile model still demonstrated statistically significant benefits in survival for the patient population selected by the model who received the treatment vs those who did not receive it. This supported that the adjustment of confounding variables was not adequate to detect the  efficacy signal by the TE-ML model and that the results should instead be attributed to confounding factors not properly corrected by the adjustment propensity method. As the futility analysis was positive, the survival effect was attributed to the confounding variables. As a result, further experimentation was aborted and survival analysis with Lopinavir-ritonavir was disregarded.

\begin{figure}[h!]
\centering
\begin{subfigure}{.49\linewidth}
  \centering
  \label{fig:NonAdjustedGeneralCovidPopulationTocilizumab}
  \caption{Non-adjusted General Covid Population}
  \includegraphics[width=1\linewidth]{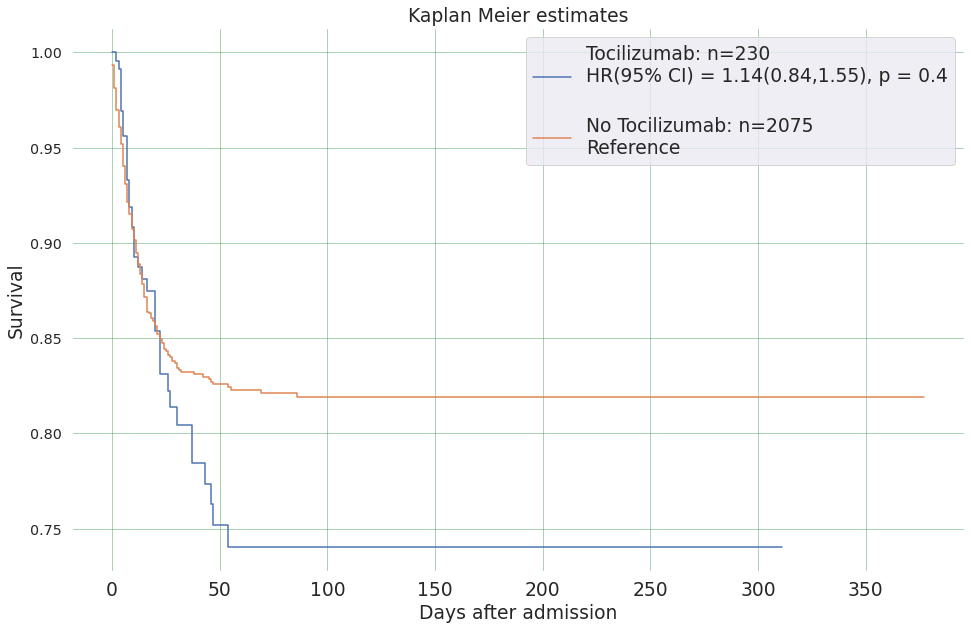}
\end{subfigure}
\begin{subfigure}{.49\linewidth}
  \centering
  \label{fig:AdjustedGeneralCovidPopulationTocilizumab}
  \caption{Adjusted General Covid Population} 
  \includegraphics[width=1\linewidth]{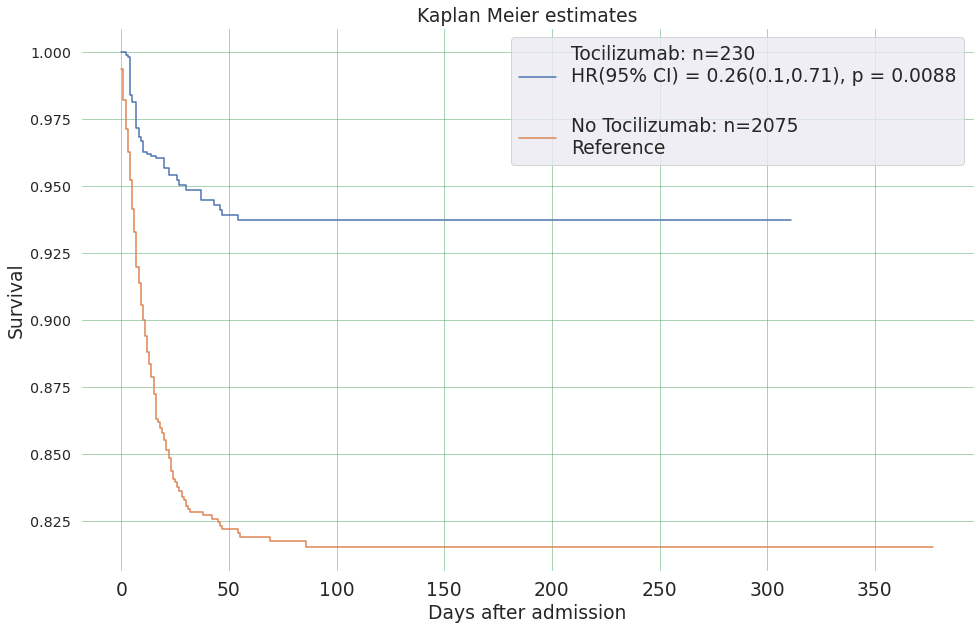}
\end{subfigure}
\begin{subfigure}{.49\textwidth}
  \centering
  \label{fig:AdjustedGuidelinesIndicatedCovidPopulationTocilizumab}
  \caption{Adjusted Guidelines Indicated} 
  \includegraphics[width=1\linewidth]{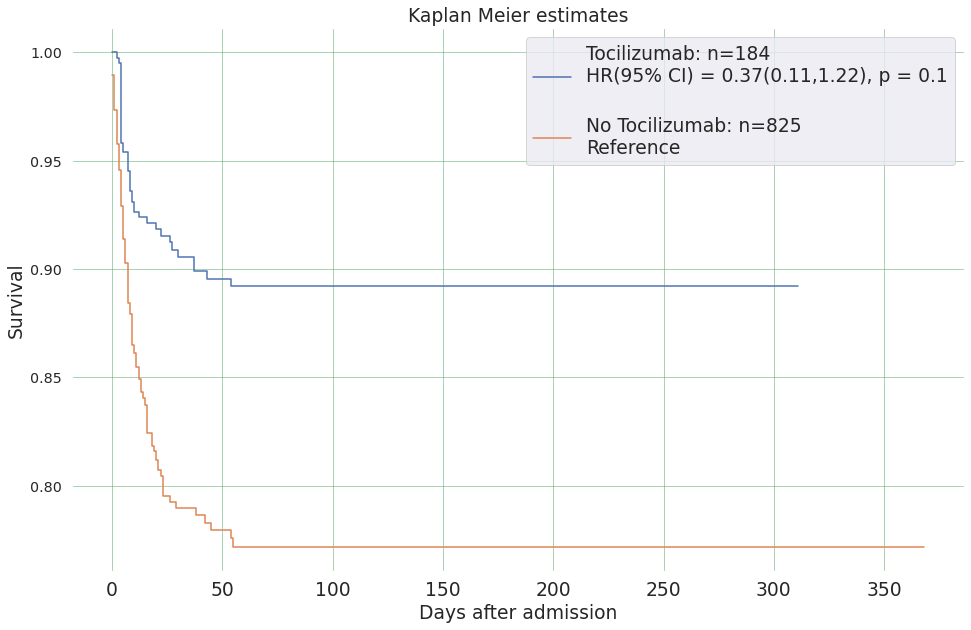}
\end{subfigure}
\begin{subfigure}{.49\textwidth}
  \centering
  \label{fig:AdjustedMLIndicatedCovidPopulationTocilizumab}
  \caption{Adjusted ML Indicated} 
  \includegraphics[width=1\linewidth]{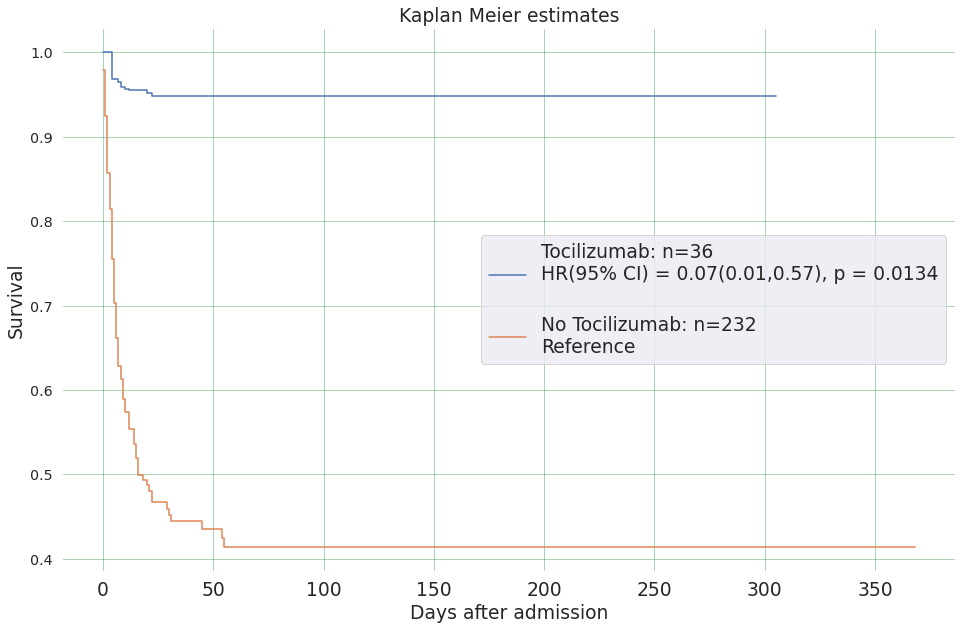}
\end{subfigure}
\caption{\textbf{Tocilizumab}: Survival analysis in the test population.a) Non-adjusted survival curves in treated vs non treated test population.  b) Adjusted survival curves in treated vs non treated test population. c) Adjusted survival curves in treated vs non treated test population who were indicated for treatment by clinical guidelines. d) Adjusted survival curves in treated vs non treated test population who were indicated by the ML model. e) Adjusted survival curves in treated vs non treated test population who were not indicated for treatment by the ML model.}
\label{fig:Tocilizumab}
\end{figure}

\paragraph{Tocilizumab}
Treated population vs the general COVID-19 population was younger (66.2 vs 66.11 years, with higher estimated Charlson 10 year survival (64.4 vs 62.7 years), lower day 1 SatO2 (91 vs 90\%, higher day 1 FiO2 (36\% vs 31\%) and higher all-cause mortality rate (19\% vs 16\%).
In the case of tocilizumab (see Fig~\ref{fig:Tocilizumab}, the unadjusted time-to-event analysis did not find any association of tocilizumab use with survival time, and survival curves of both treated and not treated patients overlapped. Nonetheless, after adjustment for confounding, a relationship was discovered
where tocilizumab was significantly associated with an increase in survival time in the general unselected population (HR = 0.26 , P = 0.008). The adjusted association in
the group that received supplemental oxygen was
not significant (HR = 0.37 , P = 0.1) but the protective effect was strongly pronounced among those indicated by the algorithm as suitable for tocilizumab with a statistically significant increase in survival time (HR = 0.21 , P = 0.00115). As explained in the futility analysis test (see section \ref{sec:futility}), a propensity adjusted ML futile model was trained in a dummy outcome reaching an AUC = 0.48 and contrary as for lopinavir-ritonavir, this association was not longer statistically significant in terms of benefits in survival for the patient population selected by the futile model who received the treatment vs those selected by the futile model who did not receive it. This supports that the adjustment of confounding variables was adequate to detect the treatment efficacy signal over the effects of confounding factors and hence further experiments on survival analysis could be trusted. The TE-ML model was capable of identifying high-risk, fragile patients (see Table~\ref{tab:desc_tests_results}), in whom Tocilizumab use is associated with improved survival outcomes. On the other hand, for those patients not recommended by the TE-ML model, who overall were characterized by having a better expected 10-year survival by Charlson index than those recommended by the ML model, there was still significant survival benefit from the use of Tocilizumab. In essence, there were survival benefits for a subset of patients treated with tocilizumab even if not selected by the TE-ML model.

\subsection{Clinical Characteristics of ML Treatment Indicated Population and Model Explainability }
Clinical characteristics of all test set versus selected test subset indicated by the ML to receive each COVID-19 treatment are shown in Table~\ref{tab:desc_tests_results}. All adjusted-ML models that, among those patients indicated by the ML model, showed a positive clinically significant survival difference between treated and non-treated arms ( Tocilizumab and Remdesivir), selected patients characterized by having more commorbidities, worst 10 year survival estimated by Charlson index and higher all-cause mortality than the general test population, being 50\% vs 16\% respectively.
\begin{table}[h!]
    \centering
\begin{tabular}{p{0.18\textwidth}p{0.1\textwidth}p{0.1\textwidth}p{0.1\textwidth}p{0.1\textwidth}p{0.1\textwidth}p{0.1\textwidth}p{0.1\textwidth}}
\toprule
\multicolumn{8}{c}{ML Treatment Indicated}           \\
\cmidrule(r){3-8}
&All Patients (Test)& Lopinavir-Ritonavir & Corticos-teroids & Tocilizumab & Chloroquine or Hydroxychloroquine & Azithromycin & Remdesivir \\
& (N=2390) &(N=427) &(N=493) &(N=268) &(N=437) &(N=1321) &(N=319) \\
\toprule
Age(years) & 65.60 & 71.07 & 66.49 & 79.15 & 72.47 & 66.12 & 77.47 \\
Charlson est. 10-y Sv (years) & 61.70 & 51.72 & 62.83 & 29.75 & 49.54 & 64.40 & 33.43 \\
\midrule
\textbf{Medical History} &&&&&&&\\
Respiratory & 16.03\% & 19.91\% & 22.11\% & 25.00\% & 17.16\% & 15.22\% & 24\% \\
Cardiovascular & 23.18\% & 26.00\% & 23.73\% & 35.07\% & 29.52\% & 20.89\% & 34\% \\
Active Malignancy & 4.02\% & 4.22\% & 2.64\% & 8.21\% & 6.41\% & 3.03\% & 7\% \\
\midrule
\textbf{Basal}&&&&&&&\\
SatO2 (\%) & 93.72 & 91.95 & 91.03 & 88.49 & 92.66 & 93.41 & 88.27 \\
FiO2 (\%) & 23.60 & 26.05 & 25.17 & 29.76 & 26.34 & 23.98 & 27.84 \\
RR (breaths/min) & 21.75 & 27.08 & 22.49 & 24.86 & 26.51 & 22.07 & 24.65 \\
HR (beat/min) & 93.77 & 94.46 & 97.40 & 92.55 & 92.25 & 94.17 & 94.85 \\
ANC ($10^9/L$) & 5.70 & 6.31 & 6.33 & 7.38 & 6.27 & 5.64 & 7.20 \\
Temperature (Celsius) & 36.66 & 36.77 & 36.82 & 36.54 & 36.69 & 36.70 & 36.72 \\
\midrule
\textbf{Day 1}&&&&&&&\\
SatO2 (\%) & 91.99 & 89.98 & 89.27 & 86.16 & 90.56 & 91.66 & 85.43 \\
FiO2(\%) & 31.33 & 38.14 & 36.49 & 51.13 & 36.56 & 32.11 & 47.49 \\
RR (breaths/min) & 24.03 & 28.54 & 25.14 & 27.27 & 28.33 & 24.25 & 27.22 \\
HR (beat/min) & 97.84 & 98.35 & 100.76 & 98.89 & 96.88 & 97.77 & 100.02 \\
ANC ($10^9/L$) & 5.78 & 6.33 & 6.39 & 7.34 & 6.30 & 5.73 & 7.22 \\
Temperature (Celsius) & 37.18 & 37.33 & 37.36 & 37.16 & 37.22 & 37.24 & 37.22 \\
\midrule
\textbf{Outcome}&&&&&&&\\
All-cause Mortality & 16.11\% & 29.74\% & 21.50\% & 50.75\% & 29.75\% & 15.37\% & 50\% \\
\bottomrule
\end{tabular}

\caption{Descriptive characteristics of all test set used for evaluating machine-learning treatment effect models and the descriptive characteristics of the selected subset indicated by the ML to receive each COVID-19 treatment. ANC = Absolute Neutrophil Count; RR = Respiratory Rate; HR = Heart Rate; SatO2 = peripheral Oxygen Saturation}
\label{tab:desc_tests_results}
\end{table}

\begin{figure}[h!]
\begin{subfigure}{.49\linewidth}
    \centering
    \includegraphics[width=1\linewidth]{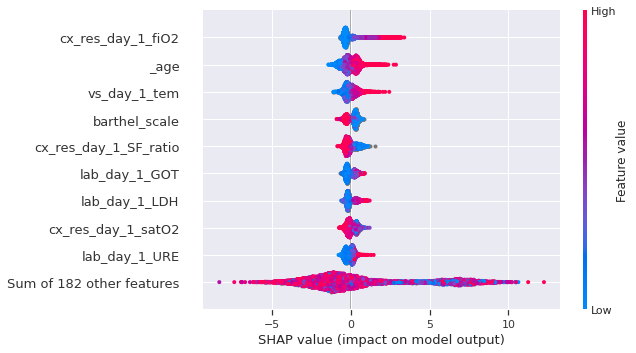}
    \caption{Tocilizumab}
\end{subfigure}
\begin{subfigure}{.49\linewidth}
    \centering
    \includegraphics[width=1\linewidth]{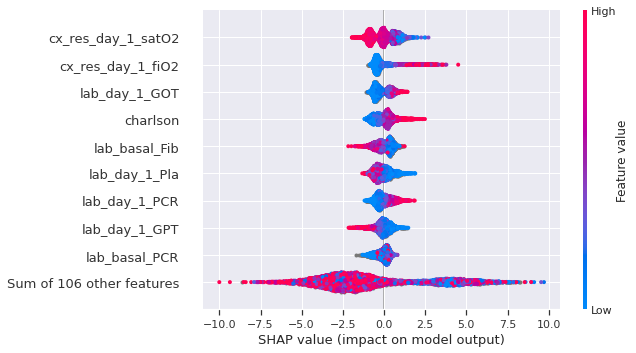}
    \caption{Remdesivir}
\end{subfigure}
\caption{\textbf{SHAP} Summary Plots for the top features for adjusted treatment-effect models for Tocilizumab and Remdesivir. Each summary plot combines feature importance with feature effects. Each point on the summary plot is a Shapley value for a feature and an instance. The position on the y-axis is determined by the feature and on the x-axis by the Shapley value. The color represents the value of the feature from low to high. Overlapping points are jittered in y-axis direction, reflecting the distribution of the Shapley values per feature. The features are ordered according to their importance.}
    \label{fig:Feature_importances}
\end{figure}

\begin{figure}[h!]
\begin{subfigure}{0.5\linewidth}
    \centering
    \includegraphics[width=1\linewidth]{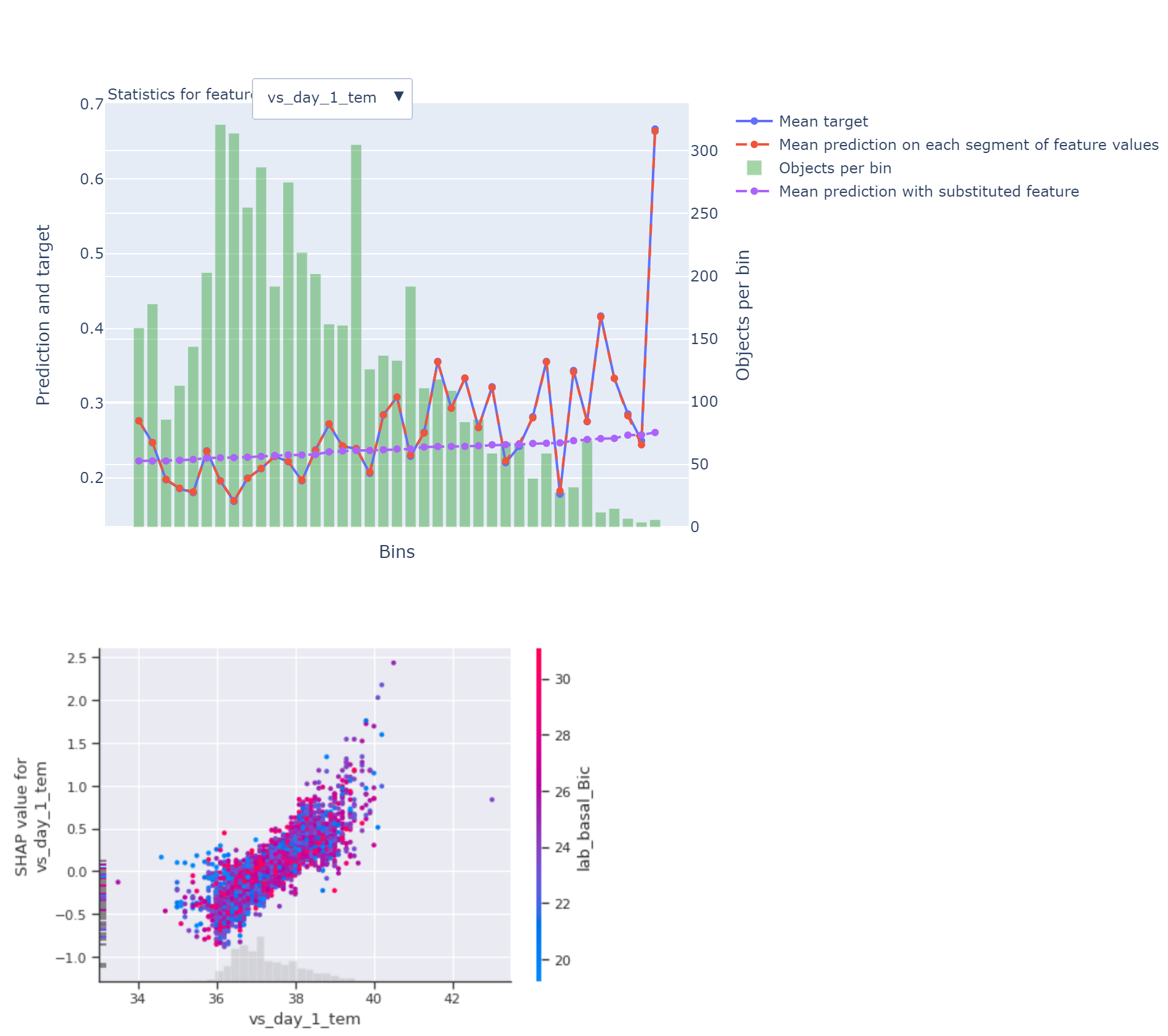}
    \caption{Temperature at day 1}
\end{subfigure}
\begin{subfigure}{0.5\linewidth}
    \centering
    \includegraphics[width=1\linewidth]{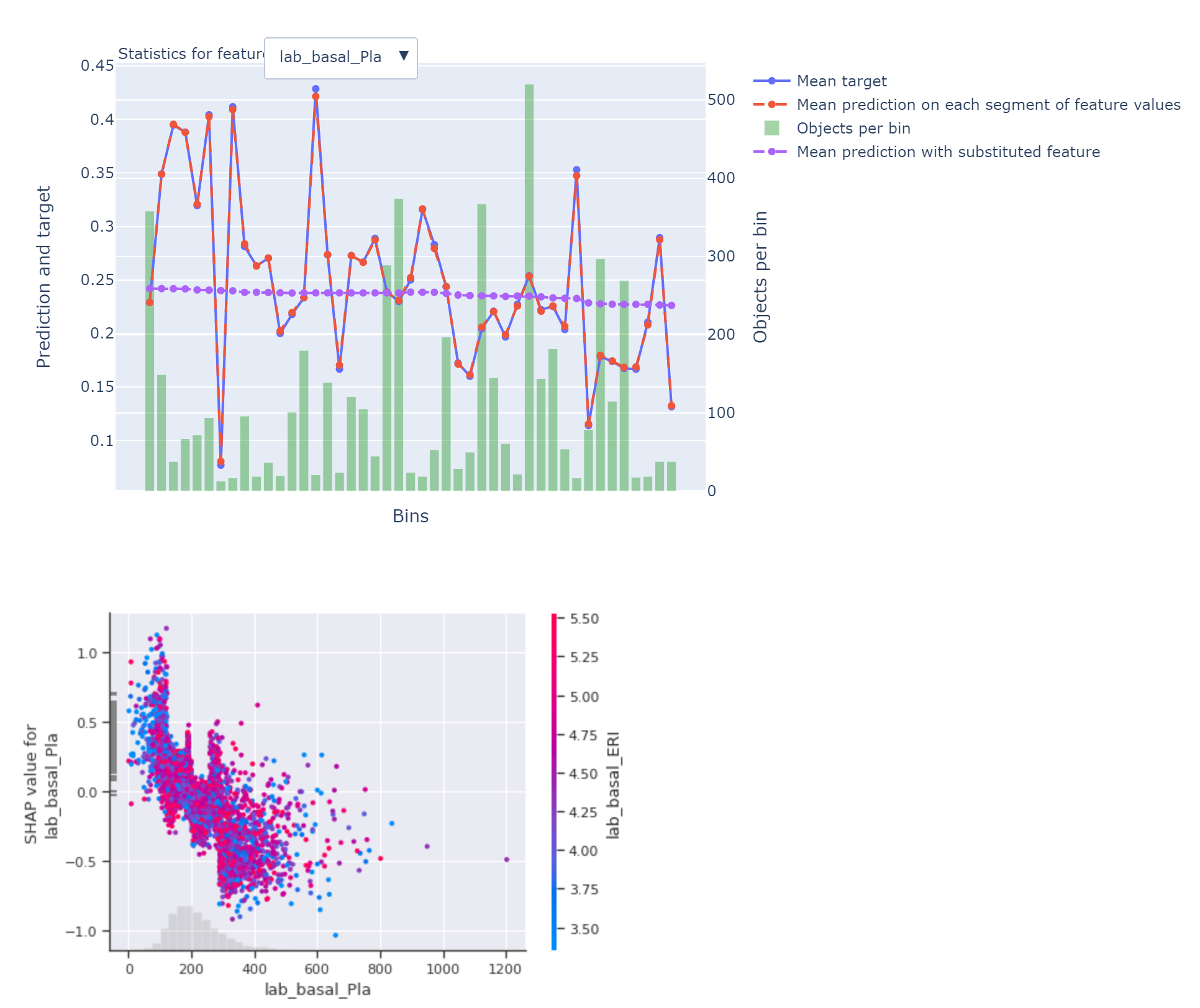}
    \caption{Platelets at day 1}
\end{subfigure}
\caption{SHAP feature dependence plots with interaction visualization for temperature and platelets at day 1}
    \label{fig:Feature_association}
\end{figure}

Most important model features for generating
predictions for a \textit{positive class} (improved disease if
treated vs worsened disease if not treated) included high FiO2 at day 1, old age and high temperature at day 1 for tocilizumab and low SatO2, high FiO2 and increased levels of GOT at day 1 for remdesivir (see Fig~\ref{fig:Feature_importances})
For features selected as most important it is also useful to explore the strength and the direction of association with the model outputs. Fig.~\ref{fig:Feature_association}
illustrates that there is a direct association between temperature and predicted benefit of tocilizumab and an indirect association with platelets where lower levels of platelets are associated with larger benefits.

\begin{figure}[h!]

    \begin{subfigure}{.49\linewidth}
    \centering
    \includegraphics[width=1\linewidth]{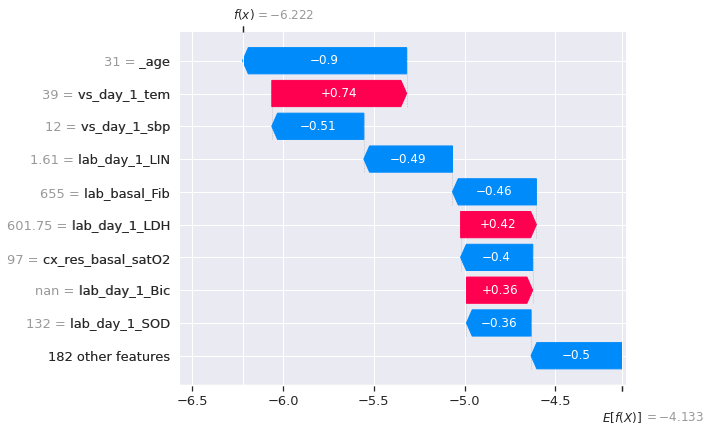}
    \caption{31 years old man}
\end{subfigure}
\begin{subfigure}{.49\linewidth}
    \centering
    \includegraphics[width=1\linewidth]{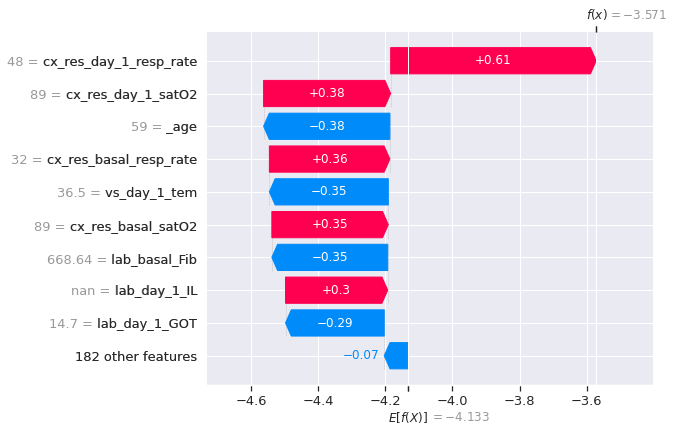}
    \caption{58 years old man}
\end{subfigure}
\caption{Treatment-effect model predictions at patient level for tocilizumab indication}
\label{fig:examples_patient}
\end{figure}

Tools used to explain ML models outputs, also allow to understand predictions at case level. Given a particular patient with a set of specific values for each of the clinical variables inputed in the model, it is possible to explore which are the most decisive feature values and on which direction they influenced the model prediction. For the Tocilizumab ML model, the average treatment benefit in the whole study dataset was -4.1 (base value) in a range between minimum -10 up to maximum recommendation of 10. The negative value is explained in part as only 9.4\% patients received the treatment.  As shown in Fig.~\ref{fig:examples_patient}    a 31 years-old male patient positive for COVID-19 with 100\% punctuation in the Barthel scale, an expected 10-years Charlson survival of 98\%, radiological findings of covid pneumonia, not supplemental oxygen with a SatO2/FiO2 ratio = 462, basal SatO2 = 97 and fever of 39º at day 1, the model outputs -6.2, shifting the base value to a more negative value or more pronounced lack of benefit from tocilizumab. On the contrary as shown in Fig.~\ref{fig:examples_patient}.b a 58 years-old man positive for COVID-19 with an expected 10-years Charlson survival of 95\%, radiological findings of covid pneumonia, a SatO2/FiO2 ratio = 241, basal SatO2 = 89, respiratory rate of 32 and 48 breaths/minute basal and at day 1 respectively, temperature of 36.5º at day 1 the model predicts a benefit gain of treatment with Tocilizumab and shifts the output towards the right.

\section{Discussion}
This study provides further evidence that ML approaches are capable of identifying which particular patients would benefit from COVID-19 pharmacotherapy based on EHR data. 

For Remdesivir, and Tocilizumab, in an independent test population, the ML models identified the groups of patients where there was a statistically significant survival benefit, comparing retrospectively those who were treated vs those not treated. The magnitude of this benefit was more pronounced than that observed in the unselected test population and in those selected based on the requirement of supplemental oxygen respectively as dictated by clinical guidelines. This supports that ML based approaches would enable precision medicine for a more personalized and rational prescription being potentially superior to current clinical guidelines and/or clinical practice.
The results obtained for remdesivir and tocilizumab are also aligned with clinical trials evidence. Specifically for Tocilizumab, in two open-label trials that included patients on oxygen support with a C-reactive protein level $>=75$ mg/L or patients who had recently started high-flow oxygen or more intensive respiratory support, adding tocilizumab reduced 28-day mortality (28 to 29 percent versus 33 to 36 percent with usual care alone) \citep{remap2021,horby2021}. In the case of remdesivir, in a meta-analysis of four trials that included over 7000 patients with COVID-19, remdesivir did not reduce mortality (OR 0.9, 95\% CI 0.7-1.12) or need for mechanical ventilation (OR 0.90, 95\% CI 0.76-1.03) compared with standard of care or placebo \cite{siemieniuk2020}. This analysis, however, grouped patients with COVID-19 of all severities together, and based on results from one included placebo-controlled trial, there may be a mortality benefit for select patients with severe disease who only require low-flow supplemental oxygen. 

Equally relevant is to be able to prove that the proposed ML approach effectively identifies patients not candidate for treatment by doing a reciprocal analysis - referred as negative treatment effect survival analysis -  to test the lack or even detrimental effect of treatment on patients who were not recommended by the ML model. The negative treatment effect survival analysis showed that the ML model effectively identified patients who would not derive survival benefits from either remdesivir or corticosteroids in the test population. While the negative results for remdesivir or corticosteroids did not reach statistical significance, the protective effect in the non-selected population clearly vanished with an inversion of the survival curves and a HR >= 1.2. For tocilizumab, the ML models failed to identify patients that are not candidate for those treatments in the test population, supporting the hypothesis that the TE-ML models only were able to identify a subset of the population who benefits, but not all. As shown in the adjusted survival analysis for unselected test population, their administration was associated with a strong beneficial effect on survival, almost comparable to those obtained in the population selected by the ML models. This observation together with the fact that the populations indicated by the ML models were enrichted with worst all-cause mortality, 50\% vs 16.1\% for the unselected test population indicates that there was a subset of less critical patients that also benefits from those treatments but were not identified by the ML models. In general the prescription at the discretion of physicians (general unselected test population) proved to be almost as effective in terms of patient selection for survival gains than the ML-guided prescription, being both statistically significant.

In line with the published scientific evidence for lack of efficacy for chloroquine or hydroxicholoroquine \citep{cavalcanti2020hydroxychloroquine}, and for azithromycin \citep{abaleke2021azithromycin}, the TE-ML models on the present study achieved a low performance both in validation and test sets, which, as reflected in the corresponding survival analysis tests, failed to select candidate patients associated with statistically significant gains in survival times. 

In the case of Lopinavir-ritonovar there was statistically significant time gains in the survival analysis in the test set but the result was not trusted because the additive analysis proved that the adjustment of confounding variables was sub-optimal. Our results hence could not conclude any positive effect. Randomized trials have not demonstrated any clinical benefit from lopinavir-ritonavir \citep{horby2020lopinavir}.

Strengths of this study:
Covering 12 different health-department provides an opportunity to capture larger data variability and to analyse patients across different demographics which strongly support that our data-driven approach rely on data that truly represent the underlying data distribution of the problem. While this is a well-known requirement, state-of-the-art algorithms are usually evaluated on carefully curated data sets, often originating from only a few sources. This can introduce biases where demographics (e.g., gender, age) or technical imbalances (e.g., acquisition protocol, equipment manufacturer) skew predictions and adversely affect the accuracy for certain groups or sites. However, to capture subtle relationships between disease patterns, demographic factors, as well as complex and rare cases, it is crucial to expose a model to diverse cases and clinical settings that largely differed across hospitals during the pandemic. In addition reserving 2 entire health departments as independent population test allow evaluate the ability of the treatment-effect models to generalize to different hospitals. In this study, all models performed worst in the test set as compared to the validation set. For remdesivir and tocilizumab, the only models where survival analysis yielded statistically significant results in the test set and were trusted after the futility analysis, the AUC degraded from 0.83 to 0.74 from validation to test set. 
Nonetheless, the ML model performance for Remdesivir was higher than the previously reported AUC of 0.57 \citep{LAM2021}, even if the AUCs obtained on the present study, 0.83 and 0.73 in validation and test set respectively, still did not reach the commonly accepted threshold of an AUC > 0.85 considered as reasonable decision making. For both drugs studied by \cite{LAM2021}, Remdesivir and Corticosteroids the direction and magnitude of treatment effects as well as the behaviour after model adjustment was similar, strengthening the evidence for the robustness of this ML-approach as well as the replicability of results. Additionally on this study we explored the negative treatment effect, or lack of benefit, and showed to be reciprocal to the positive treatment effect for both remdesivir and corticosteroids, supporting that in the population not recommended to treatment, those that actually received them retrospectively did not derive any benefits in terms of survival compared to those that did not receive them.
Another strength of this study is that its design assumes that residual confounding still exists even if a propensity- score adjustment has been done based on pretreatment confounders. As previously acknowledged by \cite{LAM2021}, it is possible that residual confounding may have influenced the results despite the efforts to adjust for confounding variables. Under this premise, the futility analysis aims to increase the confidence in the treatment-effect estimator by discarding TE-ML models that do not have additive predictive capabilities to detect the efficacy signal over residual confounding. For those cases either the treatment effect is non-existent and only confounders determines the outcome or conversely there is a treatment effect but the adjustment failed to achieve a pseudo-population effectively randomized for pre-treatment confounders among treated and not treated patients. On this study this method leads us to not trust the survival analysis for lopinavir-ritonavir.

Lastly this study exemplifies how available ML tools for explainability could be potentially be used at patient-level for a more tailored and informed decision in the clinical practice. ML-based approaches accept larger set of clinical variables that better represent each patient profile as opposed to the reduced set of selected variables used by clinical guidelines. While none of the trained models surpassed the accepted threshold of 0.85 to be applicable to the real practice, we illustrate with an example how those models could be used hypothetically to guide treatment decisions at the patient level.

Potential limitations in this study are as follows:
Not all health departments contributed all admissions attended during the study period, but as it obeyed to administrative reasons and not to medical reasons, it is not expected to have introduced a systemic selection bias in the population studied.
Each admission was highly dimensional, with number of features approx. 1950, highly sparse as many laboratory test are not done by routine to most patients and completeness was low. 
As opposed to clinical trials, our data was not manually curated at patient level, and was supervised in an aggregated manner using descriptive statistics and exploratory data analysis. The hypothesis is that at larger amount of data, algorithms are more robust to data errors, inaccuracies and missings.
Also, the small sample size for each treatment precluded any analysis of
combinatorial treatments. 
Another limitation is that this study focused only on analyzing an efficacy outcome based on survival, and was not designed to study treatment safety and/or tolerability. Such an objective would require adverse event ascertainment from EHR data, in particular from clinical notes found in natural language which was out of the scope of the present study.
\section{Conclusions}
Machine learning methods are suitable tools toward precision medicine to prescribing COVID-19 therapies. This study replicated the method used in a prior study on corticosteroids and remdesivir and obtained concordant results. Furthermore, the method effectively identified the subgroups of patients that would derive survival benefits from the prescription of remdesivir and tocilizumab, which were statistically significant in an independent population test, helping to validate the results of randomized clinical trials. At the same time, this study also add evidence for the lack of benefit from chloroquine derivates, lopinavir-ritonavir and azithromycin.

\begin{table}[h!]
	\centering
	\begin{tabular}{llll}
		\toprule
		\multicolumn{2}{c}{UMLS Metathesaurus CUIs}                   \\
		\cmidrule(r){1-2}
		Term Hierarchy     & CUI     & Leaf entity counts & Branch entity counts \\
		\midrule
├── infiltrates &C0277877& 1392& 7166\\
│   ├── interstitial pattern &C2073538& 1549& 2721\\
│   │   ├── ground glass pattern &C3544344& 1006& 1006\\
│   │   ├── reticular interstitial pattern &&126& 126\\
│   │   ├── reticulonodular interstitial pattern &C2073672& 30& 30\\
│   │   └── miliary opacities &C2073583& 10& 10\\
│   └── alveolar pattern &C1332240& 1558& 3053\\
│       ├── consolidation &C0521530& 1405& 1450\\
│       │   └── air bronchogram &C3669021& 45& 45\\
│       └── air bronchogram &C3669021& 45& 45\\
├── increased density &C1443940& 2216& 2216\\
├── pneumonia &C0032285& 2186& 5308\\
│   └── atypical pneumonia &C1412002& 193& 3122\\
│       └── viral pneumonia &C0032310& 255& 2929\\
│           \hspace{1cm}├── COVID 19 &C5203670& 2359& 2359\\
│           \hspace{1cm}└── COVID 19 uncertain &C5203671& 315& 315\\
		\bottomrule
	\end{tabular}
	\caption{COVID-19 radiological entities: selected radiological entities, corresponding UMLS Metathesaurus unique identifiers (CUIs) and counts in study population}
	\label{tab:CUIs}
\end{table}

\bibliographystyle{unsrtnat}

\begin{thebibliography}{22}
\providecommand{\natexlab}[1]{#1}
\providecommand{\url}[1]{\texttt{#1}}
\expandafter\ifx\csname urlstyle\endcsname\relax
  \providecommand{\doi}[1]{doi: #1}\else
  \providecommand{\doi}{doi: \begingroup \urlstyle{rm}\Url}\fi

\bibitem[Sherman et~al.(2016)Sherman, Anderson, Dal~Pan, Gray, Gross, Hunter,
  LaVange, Marinac-Dabic, Marks, Robb, et~al.]{sherman2016}
Rachel~E Sherman, Steven~A Anderson, Gerald~J Dal~Pan, Gerry~W Gray, Thomas
  Gross, Nina~L Hunter, Lisa LaVange, Danica Marinac-Dabic, Peter~W Marks,
  Melissa~A Robb, et~al.
\newblock Real-world evidence—what is it and what can it tell us.
\newblock \emph{N Engl J Med}, 375\penalty0 (23):\penalty0 2293--2297, 2016.

\bibitem[Jarow et~al.(2017)Jarow, LaVange, and Woodcock]{jarow2017}
Jonathan~P Jarow, Lisa LaVange, and Janet Woodcock.
\newblock Multidimensional evidence generation and fda regulatory decision
  making: defining and using “real-world” data.
\newblock \emph{Jama}, 318\penalty0 (8):\penalty0 703--704, 2017.

\bibitem[Franklin et~al.(2021)Franklin, Lin, Gatto, Rassen, Glynn, and
  Schneeweiss]{franklin2021}
Jessica~M. Franklin, Kueiyu~Joshua Lin, Nicolle~M. Gatto, Jeremy~A. Rassen,
  Robert~J. Glynn, and Sebastian Schneeweiss.
\newblock Real-world evidence for assessing pharmaceutical treatments in the
  context of covid-19.
\newblock \emph{Clinical Pharmacology \& Therapeutics}, 109\penalty0
  (4):\penalty0 816--828, 2021.
\newblock \doi{https://doi.org/10.1002/cpt.2185}.
\newblock URL
  \url{https://ascpt.onlinelibrary.wiley.com/doi/abs/10.1002/cpt.2185}.

\bibitem[Huang et~al.(2018)Huang, Clayton, Matyunina, McDonald, Benigno,
  Vannberg, and McDonald]{huang2018}
Cai Huang, Evan~A Clayton, Lilya~V Matyunina, L~DeEtte McDonald, Benedict~B
  Benigno, Fredrik Vannberg, and John~F McDonald.
\newblock Machine learning predicts individual cancer patient responses to
  therapeutic drugs with high accuracy.
\newblock \emph{Scientific reports}, 8\penalty0 (1):\penalty0 1--8, 2018.

\bibitem[Lam et~al.(2021)Lam, Siefkas, Zelin, Barnes, Dellinger, Vincent,
  Braden, Burdick, Hoffman, Calvert, Mao, and Das]{LAM2021}
Carson Lam, Anna Siefkas, Nicole~S. Zelin, Gina Barnes, R.~Phillip Dellinger,
  Jean-Louis Vincent, Gregory Braden, Hoyt Burdick, Jana Hoffman, Jacob
  Calvert, Qingqing Mao, and Ritankar Das.
\newblock Machine learning as a precision-medicine approach to prescribing
  covid-19 pharmacotherapy with remdesivir or corticosteroids.
\newblock \emph{Clinical Therapeutics}, 2021.
\newblock ISSN 0149-2918.
\newblock \doi{https://doi.org/10.1016/j.clinthera.2021.03.016}.
\newblock URL
  \url{https://www.sciencedirect.com/science/article/pii/S0149291821001284}.

\bibitem[Bustos et~al.(2020)Bustos, Pertusa, Salinas, and {de la
  Iglesia-Vayá}]{Bustos2020}
Aurelia Bustos, Antonio Pertusa, Jose-Maria Salinas, and Maria {de la
  Iglesia-Vayá}.
\newblock Padchest: A large chest x-ray image dataset with multi-label
  annotated reports.
\newblock \emph{Medical Image Analysis}, 66:\penalty0 101797, 2020.
\newblock ISSN 1361-8415.
\newblock \doi{https://doi.org/10.1016/j.media.2020.101797}.
\newblock URL
  \url{https://www.sciencedirect.com/science/article/pii/S1361841520301614}.

\bibitem[de~la Iglesia~Vayá et~al.(2020)de~la Iglesia~Vayá, Saborit, Montell,
  Pertusa, Bustos, Cazorla, Galant, Barber, Orozco-Beltrán, García-García,
  Caparrós, González, and Salinas]{Vaya2020}
Maria de~la Iglesia~Vayá, Jose~Manuel Saborit, Joaquim~Angel Montell, Antonio
  Pertusa, Aurelia Bustos, Miguel Cazorla, Joaquin Galant, Xavier Barber,
  Domingo Orozco-Beltrán, Francisco García-García, Marisa Caparrós, Germán
  González, and Jose~María Salinas.
\newblock Bimcv covid-19+: a large annotated dataset of rx and ct images from
  covid-19 patients, 2020.

\bibitem[Cao et~al.(2020)Cao, Wang, and et~al.]{Cao2020}
Bin Cao, Yeming Wang, and Wen et~al.
\newblock A trial of lopinavir–ritonavir in adults hospitalized with severe
  covid-19.
\newblock \emph{New England Journal of Medicine}, 382\penalty0 (19):\penalty0
  1787--1799, 2020.
\newblock \doi{10.1056/NEJMoa2001282}.
\newblock URL \url{https://doi.org/10.1056/NEJMoa2001282}.
\newblock PMID: 32187464.

\bibitem[O’Kelly and Li(2020)]{Kelly2020}
Michael O’Kelly and Siying Li.
\newblock Assessing via simulation the operating characteristics of the who
  scale for covid-19 endpoints.
\newblock \emph{Statistics in Biopharmaceutical Research}, 12\penalty0
  (4):\penalty0 451--460, 2020.
\newblock \doi{10.1080/19466315.2020.1811148}.
\newblock URL \url{https://doi.org/10.1080/19466315.2020.1811148}.

\bibitem[Haimovich et~al.(2020)Haimovich, Ravindra, Stoytchev, Young, Wilson,
  van Dijk, Schulz, and Taylor]{Haimovich2020}
Adrian Haimovich, Neal~G. Ravindra, Stoytcho Stoytchev, H.~Patrick Young,
  Francis~Perry Wilson, David van Dijk, Wade~L. Schulz, and R.~Andrew Taylor.
\newblock Development and validation of the covid-19 severity index (csi): a
  prognostic tool for early respiratory decompensation.
\newblock \emph{medRxiv}, 2020.
\newblock \doi{10.1101/2020.05.07.20094573}.
\newblock URL
  \url{https://www.medrxiv.org/content/early/2020/05/14/2020.05.07.20094573}.

\bibitem[Linssen et~al.(2020)Linssen, Ermens, Berrevoets, Seghezzi, Previtali,
  Russcher, Verbon, Gillis, Riedl, de~Jongh, et~al.]{linssen2020}
Joachim Linssen, Anthony Ermens, Marvin Berrevoets, Michela Seghezzi, Giulia
  Previtali, Henk Russcher, Annelies Verbon, Judith Gillis, J{\"u}rgen Riedl,
  Eva de~Jongh, et~al.
\newblock A novel haemocytometric covid-19 prognostic score developed and
  validated in an observational multicentre european hospital-based study.
\newblock \emph{Elife}, 9:\penalty0 e63195, 2020.

\bibitem[Charlson et~al.(1987)Charlson, Pompei, Ales, and
  MacKenzie]{charlson1987new}
Mary~E Charlson, Peter Pompei, Kathy~L Ales, and C~Ronald MacKenzie.
\newblock A new method of classifying prognostic comorbidity in longitudinal
  studies: development and validation.
\newblock \emph{Journal of chronic diseases}, 40\penalty0 (5):\penalty0
  373--383, 1987.

\bibitem[Xu et~al.(2010)Xu, Ross, Raebel, Shetterly, Blanchette, and
  Smith]{xu2010use}
Stanley Xu, Colleen Ross, Marsha~A Raebel, Susan Shetterly, Christopher
  Blanchette, and David Smith.
\newblock Use of stabilized inverse propensity scores as weights to directly
  estimate relative risk and its confidence intervals.
\newblock \emph{Value in Health}, 13\penalty0 (2):\penalty0 273--277, 2010.

\bibitem[Dorogush et~al.(2017)Dorogush, Gulin, Gusev, Kazeev, Prokhorenkova,
  and Vorobev]{catboost}
Anna~Veronika Dorogush, Andrey Gulin, Gleb Gusev, Nikita Kazeev,
  Liudmila~Ostroumova Prokhorenkova, and Aleksandr Vorobev.
\newblock Fighting biases with dynamic boosting.
\newblock \emph{CoRR}, abs/1706.09516, 2017.
\newblock URL \url{http://arxiv.org/abs/1706.09516}.

\bibitem[Lundberg and Lee(2017)]{lundberg2017}
Scott~M Lundberg and Su-In Lee.
\newblock A unified approach to interpreting model predictions.
\newblock In \emph{Proceedings of the 31st international conference on neural
  information processing systems}, pages 4768--4777, 2017.

\bibitem[NIH()]{NIH}
Covid-19 treatment guidelines panel. coronavirus disease 2019 (covid-19)
  treatment guidelines. national institutes of health.
\newblock \url{https://www.covid19treatmentguidelines.nih.gov/}.
\newblock Accessed: July 2021.

\bibitem[Investigators(2021)]{remap2021}
REMAP-CAP Investigators.
\newblock Interleukin-6 receptor antagonists in critically ill patients with
  covid-19.
\newblock \emph{New England Journal of Medicine}, 384\penalty0 (16):\penalty0
  1491--1502, 2021.

\bibitem[Horby et~al.(2021)Horby, Pessoa-Amorim, Peto, et~al.]{horby2021}
PW~Horby, G~Pessoa-Amorim, L~Peto, et~al.
\newblock for the recovery collaborative group. tocilizumab in patients
  admitted to hospital with covid-19 (recovery): preliminary results of a
  randomised, controlled, open-label, platform trial.
\newblock \emph{medRxiv}, pages 12--24, 2021.

\bibitem[Siemieniuk et~al.(2020)Siemieniuk, Bartoszko, Ge, Zeraatkar, Izcovich,
  Kum, Pardo-Hernandez, Rochwerg, Lamontagne, Han, et~al.]{siemieniuk2020}
Reed~Ac Siemieniuk, Jessica~J Bartoszko, Long Ge, Dena Zeraatkar, Ariel
  Izcovich, Elena Kum, Hector Pardo-Hernandez, Bram Rochwerg, Francois
  Lamontagne, Mi~Ah Han, et~al.
\newblock Drug treatments for covid-19: living systematic review and network
  meta-analysis.
\newblock \emph{Bmj}, 370, 2020.

\bibitem[Cavalcanti et~al.(2020)Cavalcanti, Zampieri, Rosa, Azevedo, Veiga,
  Avezum, Damiani, Marcadenti, Kawano-Dourado, Lisboa,
  et~al.]{cavalcanti2020hydroxychloroquine}
Alexandre~B Cavalcanti, Fernando~G Zampieri, Regis~G Rosa, Luciano~CP Azevedo,
  Viviane~C Veiga, Alvaro Avezum, Lucas~P Damiani, Aline Marcadenti,
  Let{\'\i}cia Kawano-Dourado, Thiago Lisboa, et~al.
\newblock Hydroxychloroquine with or without azithromycin in mild-to-moderate
  covid-19.
\newblock \emph{New England Journal of Medicine}, 383\penalty0 (21):\penalty0
  2041--2052, 2020.

\bibitem[Abaleke et~al.(2021)Abaleke, Abbas, Abbasi, Abbott, Abdelaziz,
  Abdelbadiee, Abdelfattah, Abdul, Rasheed, Abdul-Kadir,
  et~al.]{abaleke2021azithromycin}
Eugenia Abaleke, Mustafa Abbas, Sadia Abbasi, Alfie Abbott, Ashraf Abdelaziz,
  Sherif Abdelbadiee, Mohamed Abdelfattah, Basir Abdul, Althaf~Abdul Rasheed,
  Rezan Abdul-Kadir, et~al.
\newblock Azithromycin in patients admitted to hospital with covid-19
  (recovery): a randomised, controlled, open-label, platform trial.
\newblock \emph{The Lancet}, 397\penalty0 (10274):\penalty0 605--612, 2021.

\bibitem[Horby et~al.(2020)Horby, Mafham, Bell, Linsell, Staplin, Emberson,
  Palfreeman, Raw, Elmahi, Prudon, et~al.]{horby2020lopinavir}
Peter~W Horby, Marion Mafham, Jennifer~L Bell, Louise Linsell, Natalie Staplin,
  Jonathan Emberson, Adrian Palfreeman, Jason Raw, Einas Elmahi, Benjamin
  Prudon, et~al.
\newblock Lopinavir--ritonavir in patients admitted to hospital with covid-19
  (recovery): a randomised, controlled, open-label, platform trial.
\newblock \emph{The Lancet}, 396\penalty0 (10259):\penalty0 1345--1352, 2020.

\end{thebibliography}

\appendix
\begin{table}
\begin{tabular}{p{0.2\textwidth}p{0.3\textwidth}p{0.3\textwidth}p{0.1\textwidth}}
\toprule
Classification & Variable & Type & Unit \\
\midrule

\multirow{3}{3cm}{Demographic} & Health department$^{c}$ & Categorical & \\
& Age$^{c,i}$ & Integer & Years \\
& Gender$^{c,i}$ & Categorical & \\
\midrule
\multirow{6}{3cm}{Admission characteristics} & Date of hospital admission$^{c}$ & Date & \\
& Length of stay & Integer & Days \\
& Length of ICU stay & Integer & Days \\
& Discharge destination & Categorical & \\
& Time from hospital admission to ICU admission & Integer if admitted to ICU, else empty & Days \\
& Time from hospital discharge to readmission & Integer if readmission, else empty & Days \\
\midrule
\multirow{11}{3cm}{Commorbidities and Risk Factors} & Lipidemias$^{c,i}$  & Binary & \\
& Nicotine dependence and Tobacco use$^{c,i}$  & Binary & \\
& High Blood Pressure$^{c,i}$  & Binary & \\
& Diabetes mellitus type 1$^{c,i}$  & Binary & \\
& Diabetes mellitus type 2$^{c,i}$  & Binary & \\
& Heart chronic disease$^{c,i}$  & Binary & \\
& Immunocompromised$^{c,i}$  & Binary & \\
& Liver chronic disease$^{c,i}$  & Binary & \\
& Renal chronic disease$^{c,i}$  & Binary & \\
& Respiratory chronic disease$^{c,i}$  & Binary & \\
& Active malignancy$^{c,i}$  & Binary & \\
\midrule
\multirow{3}{3cm}{Survival} & Estimated 10-year survival according to Charlson comorbidity index & Integer 0-100 & percentage \\
& Time from admission to last follow-up & Integer & Days \\
& Time from admission to fatal outcome & Integer if fatal outcome, else empty & Days \\
\midrule
\multirow{4}{3cm}{ Clinical Respiratory} & Peripheral oxygen saturation$^{c,i}$  & Integer array that includes baseline value followed by the minimum value for each hospitalization day & percentage oxygen \\
& FiO2$^{c,i}$  & Integer array that includes baseline value followed by the max value for each hospitalization day & percentage \\
& SatO2/FiO2 ratio$^{c,i}$  & Integer array that includes baseline value followed by the min value for each hospitalization day & \\
& Respiratory rate$^{c,i}$  & Integer array that includes baseline value followed by the maximum value for each hospitalization day & Breaths/min \\
\midrule
\multirow{8}{3cm}{Clinical Scale} & Barthel independence scale$^{c,i}$ & Ordinal 0-100 & \\
& Sofa scale & Ordinal 0-24 & \\
& Glasgow scale & Ordinal 0-15 & \\
& Critical respiratory illness & Binary & \\
& Disease severity grade & Ordinal 1-5 & \\
& Modified WHO COVID outcomes scale & Array of ordinals (1-7 ) for all admission days & \\
& Charlson commorbidity index$^{c,i}$ & Ordinal 0-24 & \\
& Glasgow scale & Ordinal 1-15 & \\
\midrule
\end{tabular}
\caption{Study variables: Variables included as input features for the propensity score models are marked with $^{c}$. Variables included as input features for the treatment-effect models are marked with $^{i}$. For arrays with daily values, only the baseline and day 1 values are included in $^{c}$ and $^{i}$.  }
\label{tab:Variables1}
\end{table}

\begin{table}
\begin{tabular}{p{0.2\textwidth}p{0.5\textwidth}p{0.2\textwidth}p{0.1\textwidth}}
\toprule
Classification & Variable & Type & Unit \\
\midrule
\multirow{16}{3cm}{Complications} & COVID viral pneumonia & Binary & \\
& Adult respiratory distress syndrome & Binary & \\
& Acute bronchitis & Binary & \\
& Acute kidney injury & Binary & \\
& Disturbance of consciousness and cognition & Binary & \\
& Cerebrovascular ischemic disease & Binary & \\
& Cytokine release syndrome & Binary & \\
& Disseminated intravascular coagulation & Binary & \\
& Gastroenteritis & Binary & \\
& Heart disease & Binary & \\
& Systemic inflammatory response syndrome & Binary & \\
& Lung embolism & Binary & \\
& Secondary bacterial pneumonia & Binary & \\
& Deep venous thrombosis & Binary & \\
\midrule
\multirow{3}{3cm}{Image} & Number of image studies with COVID associated radiological findings & Integer & \\
& COVID associated radiological findings$^{c,i}$  & Binary & \\
& Number of image studies during hospitalization & Integer & \\
\midrule
\multirow{2}{3cm}{Laboratory} & 25-Hydroxy Vitamin D,Albumin, B-Cells, B-type natriuretic peptide, Bicarbonate, Bilirubin, Calcium, Chloride, Creatine phosphokinase, Creatinine, D dimer, Erythrocyte, Glomerular filtration rate CKD-EPI, Glomerular filtration rate MDRD-4, Phosphate, Ferritine, Fibrinogen, Gamma-glutamyl transferase, Glucose, Aspartate Aminotransferase, Alanine Aminotransferase, Immature granulocytes, Glycosylated hemoglobin, Hemoglobin, Days from admission to hypoglycemic episode, Hypoglycemic episode grade ( Ordinal 0 - 3), Interleukin, Interleukin 6, Urine potassium, Lactate dehydrogenase, Lymphocytes absolute count, Leucocytes absolute count, Magnessium, Monoytes absolute count, Neutrophiles absolute count, Neutrophil to lymphocytes ratio, C-Reactive Protein, Potassium, Protein, Plaquelets absolute count, procalcitonin, Prealbumin, Sodium, T Cell Count, CD4 Lymphocyte Count, CD8 Lymphocyte Count, Troponin, Partial thromboplastin time, Transferrin, Urea, Zinc$^{c,i}$  & Integer array that includes baseline value followed by the mean value for each hospitalization day & \\
& Glucose variation coefficient$^{c,i}$  & Integer array that includes the variation coefficient value for each hospitalization day & \\
\midrule
\multirow{8}{3cm}{Laboratory Respiratory} & Total CO2$^{c,i}$ &  & \\
& Base excess$^{c,i}$ &   & \\
& FiO2 reported in oxymetry$^{c,i}$ &  & \\
& Standard base excess$^{c,i}$ &  & \\
& Oxygen arterial saturation$^{c,i}$ &  & \\
& Partial pressure of CO2$^{c,i}$ &  & \\
& pH$^{c,i}$ &  & \\
& Partial pressure of Oxygen$^{c,i}$ & Integer array that includes baseline value followed by the mean value for each hospitalization day & \\
\midrule
\end{tabular}
\caption{Study variables (continuation)}
\label{tab:Variables2}
\end{table}

\begin{table}
\begin{tabular}{p{0.2\textwidth}p{0.4\textwidth}p{0.3\textwidth}p{0.1\textwidth}}
\toprule
Classification & Variable & Type & Unit \\
\midrule
\multirow{3}{3cm}{Microbiological} & Number of positive SARS-COV-2 RT-PCR & Integer & \\
& Microbiological culture positive$^{c,i}$ & Binary & \\
& Number of SARS-COV-2 RT-PCR & Integer & \\
\midrule
\multirow{32}{3cm}{COVID-19 Pharmacotherapy*} & & Integer array that includes daily doses at each admission day & \\
& Anakinra$^{c}$ &  & \\
& Azithromycin$^{c}$ &   & \\
& Baricitinib$^{c}$ &   & \\
& Bemiparine$^{c}$ &   & \\
& Ciclosporine$^{c}$ &   & \\
& Cloroquine$^{c}$ &   & \\
& Convalescent plasma$^{c}$ &   & \\
& Systemic corticosteroids$^{c}$ &   & \\
& Darunavir cobicistat$^{c}$ &   & \\
& Eculizumab$^{c}$ &   & \\
& Enoxaparine$^{c}$ &   & \\
& Fosamprenavir$^{c}$ &   & \\
& G-CSF$^{c}$ &   & \\
& Heparine$^{c}$ &   & \\
& Hidroxicloroquine$^{c}$ &   & \\
& Inmunoglobulins$^{c}$ &   & \\
& Interferon beta 1a$^{c}$ &   & \\
& Interferon beta 1b$^{c}$ &   & \\
& Ivermectine$^{c}$ &   & \\
& Lopinavir ritonavir$^{c}$ &   & \\
& Remdesivir$^{c}$ &   & \\
& Ruxolitinib$^{c}$ &   & \\
& Sarilumab$^{c}$ &   & \\
& Siltuximab$^{c}$ &   & \\
& Tacrolimus$^{c}$ &   & \\
& Tiamine vitamine B1$^{c}$ &   & \\
& Tocilizumab$^{c}$ &   & \\
& Tofacitinib$^{c}$ &   & \\
& Vitamine C$^{c}$ &   & \\
& Vitamine D$^{c}$ &  & \\
\midrule
\multirow{6}{3cm}{Vital signs} & Diastolic blood pressure$^{c,i}$ & Integer array that includes baseline value followed by the minimun value for each hospitalization day & \\
& Heart rate$^{c,i}$ & Integer array that includes baseline value followed by the mean value for each hospitalization day & \\
& Height$^{c,i}$ & Integer & \\
& Systolic blood pressure$^{c,i}$ & Integer array that includes baseline value followed by the minimun value for each hospitalization day & \\
& Temperature$^{c,i}$ & Integer array that includes baseline value followed by the maximum value for each hospitalization day & Celsius \\
& Weight$^{c,i}$ & Integer & Kilograms \\
\bottomrule
\end{tabular}
\caption{Study variables (continuation). * Covariates for propensity score  calculation, marked as $^{c}$ included all pharmacotherapy except for each therapy for which the propensity score was calculated.}
\label{tab:Variables3}
\end{table}

\section*{Acknowledgement}

This work was funded by the AVI (Valencian Innovation Agency) through the urgent aid program for scientific-innovative solutions to the fight against COVID-19, approved by decree 63/2020, May 15. 

We thank Antonio Pertusa, PhD from the Pattern Recognition and Artificial Intelligence Group (GRFIA) and the University Institute for Computing Research (IUII) at the University of Alicante for his review and constructive suggestions.

\end{document}